\pdfoutput=1
\documentclass[11pt]{article}
\usepackage[preprint]{acl}
\usepackage{times}
\usepackage{latexsym}
\usepackage[T1]{fontenc}
\usepackage[utf8]{inputenc}
\usepackage{microtype}
\usepackage{inconsolata}
\usepackage{graphicx}
\usepackage{multirow}
\usepackage{xcolor}
\usepackage{amsmath}
\usepackage{amssymb}
\usepackage{booktabs}
\usepackage{lipsum}
\usepackage{xcolor}
\definecolor{mygreen}{RGB}{1,113,0}
\usepackage{colortbl}
\definecolor{lightbluegray}{RGB}{235,245,255}
\definecolor{lightgreengray}{RGB}{235,245,235}
\usepackage{pifont}
\usepackage[utf8]{inputenc}
\usepackage{mdframed}
\usepackage{enumitem}
\usepackage[most]{tcolorbox}
\usepackage{algorithm,algpseudocode}
\usepackage{adjustbox}
\usepackage[utf8]{inputenc}
\usepackage[T1]{fontenc}   
\usepackage{hyperref}      
\usepackage{url}     
\usepackage{amsfonts}      
\usepackage{nicefrac}      
\usepackage{microtype} 
\usepackage{wrapfig}
\usepackage{arydshln}

\definecolor{customgreen}{RGB}{199,234,92}

\title{MemeReaCon: Probing Contextual Meme Understanding in\\Large Vision-Language Models}

\author{
  \textbf{Zhengyi Zhao\textsuperscript{1}},
  \textbf{Shubo Zhang\textsuperscript{2}},
  \textbf{Yuxi Zhang\textsuperscript{2}},
  \textbf{Yanxi Zhao\textsuperscript{2}},
  \textbf{Yifan Zhang\textsuperscript{2}},\\
  \textbf{Zezhong Wang\textsuperscript{1}},
  \textbf{Huimin Wang\textsuperscript{3}},
  \textbf{Yutian Zhao\textsuperscript{3}},
  \textbf{Bin Liang\textsuperscript{1}},
  \textbf{Yefeng Zheng\textsuperscript{4}},\\
  \textbf{Binyang Li\textsuperscript{2}},
  \textbf{Kam-Fai Wong\textsuperscript{1}},
  \textbf{Xian Wu\textsuperscript{3,\thanks{Corresponding author.}}},
\\
\\
  \textsuperscript{1} The Chinese University of Hong Kong
  \textsuperscript{2} University of International Relations \\
  \textsuperscript{3} Jarvis Research Center, Tencent YouTu Lab 
  \textsuperscript{4} Westlake University
\\
  {
   \texttt{zyzhao@se.cuhk.edu.hk}
  }
}

\begin{document}
\maketitle

\begin{abstract}
Memes have emerged as a popular form of multimodal online communication, where their interpretation heavily depends on the specific context in which they appear. Current approaches predominantly focus on isolated meme analysis, either for harmful content detection or standalone interpretation, overlooking a fundamental challenge: the same meme can express different intents depending on its conversational context. This oversight creates an evaluation gap: although humans intuitively recognize how context shapes meme interpretation, Large Vision Language Models (LVLMs) can hardly understand context-dependent meme intent. To address this critical limitation, we introduce MemeReaCon, a novel benchmark specifically designed to evaluate how LVLMs understand memes in their original context. We collected memes from five different Reddit communities, keeping each meme's image, the post text, and user comments together. We carefully labeled how the text and meme work together, what the poster intended, how the meme is structured, and how the community responded. Our tests with leading LVLMs show a clear weakness: models either fail to interpret critical information in the contexts, or overly focus on visual details while overlooking communicative purpose. MemeReaCon thus serves both as a diagnostic tool exposing current limitations and as a challenging benchmark to drive development toward more sophisticated LVLMs of the context-aware understanding.
\end{abstract}

\section{Introduction}
\begin{figure}[!t]
    \centering
    \includegraphics[width=.95\linewidth]{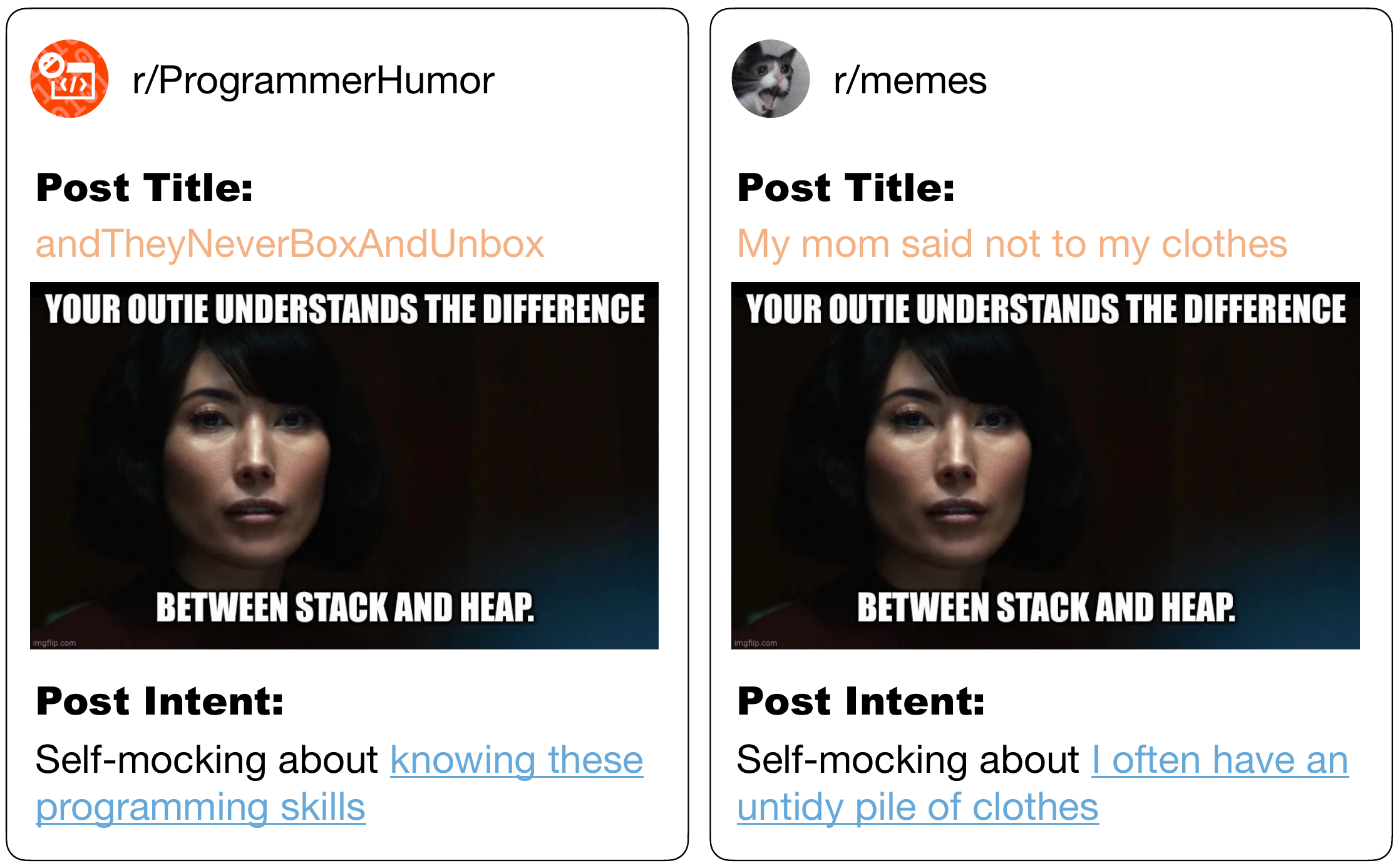}
    \caption{Demo illustrating how a single meme's interpretation changes across different contextual settings. Meme here literally indicates \textit{you know the difference between ``Stack'' and ``Heap''}. The ``Stack'' and ``Heap'' mean specific terms in programmer community, but can mean condition of an item in general talk.}
    \label{fig:demo_show_post_intent}
\end{figure}

Memes are ``amateur media artifacts, extensively remixed and recirculated by different participants on social media networks'' \cite{milner2012world} that have become a key part of how people communicate online. These combinations of images and text derive meaning not just from their content, but from their contextual placement: where they appear, why they are shared, and how communities respond to them. A meme posted in a programmer joke forum carries a fundamentally different meaning than the same meme shared in a generic community, as illustrated in Figure~\ref{fig:demo_show_post_intent}. While humans naturally process these contextual distinctions, developing computational models that can achieve similar understanding remains a significant challenge \cite{wang2024they}.

Current meme-focused research has largely pursued two distinct paths, neither fully capturing the contextual richness of memes in real online communication. The first approach centers on detecting harmful or toxic meme content \cite{sharma2022findings,hee2023decoding,huang2024towards}. While crucial for content moderation systems, this research typically leverages context primarily as a classifier for harmfulness rather than for comprehensive meaning interpretation. The second research direction tackles isolated meme understanding through tasks like caption generation \cite{hwang2023memecap}, intent description \cite{park2024memeintent}, and role explanation \cite{sharma2023you}. Despite their value, these efforts examine memes divorced from their original context, separating them from post text, creator intent, and community reactions that collectively shape their contextual meaning.

This decontextualization creates a fundamental evaluation gap: we lack methods to assess whether LVLMs can understand why particular memes are selected for specific communicative situations. As \citet{park2024memeintent} observed, people create memes ``with an intent to perform some action''. The same meme template can convey radically different meanings depending on its accompanying post title, community norms, or ongoing conversation thread \cite{lin2024goat}. Without incorporating these contextual elements, we cannot effectively measure LVLMs' capacity to process memes as humans naturally do in online environments.

To address these limitations, we developed \textbf{MemeReaCon}: Meme Reasoning in Context, a comprehensive benchmark specifically designed to evaluate LVLMs' ability to understand memes within their original contexts. We constructed MemeReaCon using content from five diverse Reddit communities, encompassing varied topics, styles, and community norms. Each example preserves three critical contextual elements: the meme image itself, the complete post text, and the top-rated community comments that reveal collective interpretation. Beyond mere data collection, our benchmark includes detailed annotations that enable targeted analysis of specific contextual understanding dimensions.

Through MemeReaCon, we investigate two fundamental questions about current LVLM limitations: (1) To what extent do models understand the meme? (2) To what extent does the post context affect models' understanding of meme?

Our extensive evaluation of leading LVLMs reveals a persistent weakness in contextual integration. Models frequently fail to establish meaningful connections between memes and their context, \textbf{either fail to interpret critical information in the contexts, or overly focus on visual details while overlooking communicative purpose.} Detailed error analysis reveals that models are sensitive to context type, such that models often fail in culturally dominant contexts rather than giving specific tags or communities. Our work makes following contributions:

\begin{itemize}[leftmargin=*, itemsep=0pt]
    \item To our knowledge, we firstly identify how the post context and meme work together: post context mainly explains the meme, or the meme illustrates points made in the context. This helps us to evaluate models whether understand different ways people use memes to communicate.
    \item We propose a novel benchmark, MemeReaCon, for meme understanding that maintains the essential relationship between meme images, post, and community reception, enabling the first systematic evaluation of how well LVLMs interpret memes as they actually function in online environments.
    \item We conduct comprehensive evaluation, revealing contextual-insensitive limitations in current LVLMs to connect multimodal elements for contextual interpretation.
\end{itemize}

\begin{table*}[!t]
\centering
\adjustbox{max width=\linewidth}{
\begin{tabular}{lllll}
\toprule
\textbf{Dataset} & \textbf{Task Type} & Post Context & Comments & Size\\
\midrule
MultiOFF \cite{suryawanshi2020multimodal}     & classify: meme hatefulness& \ding{56} & \ding{56} & 743\\
HatefulMemes \cite{kiela2020hateful} & classify: meme hatefulness& \ding{56} & \ding{56} & 10k\\
Jewtocracy \cite{chandra2021subverting}   & classify: meme hatefulness& \ding{56} & \ding{56} & 6,611\\
HarMeme \cite{pramanick2021detecting}     & classify: meme hatefulness/target& \ding{56} & \ding{56} & 3,544\\
MAMI \cite{fersini2022semeval}        & classify: meme hatefulness& \ding{56} & \ding{56} & 15k\\
FigMemes \cite{liu2022figmemes}    & classify: meme political opinion& \ding{56} & \ding{56} & 5,141\\
HVVMemes \cite{sharma2022findings}    & classify: meme character role& \ding{56} & \ding{56} & 7k\\
GOAT \cite{lin2024goat}  & classify: meme hatefulness& \ding{56} & \ding{56} & 6,626\\
\hdashline
HatReD \cite{hee2023decoding}      & explain: meme                                 & \ding{56} & \ding{56} & 3,228\\
ExHVV \cite{sharma2023you}       & explain: meme               & \ding{56} & \ding{56} & 4,680\\
MemeCap \cite{hwang2023memecap}     & explain: meme + metaphors                                   & \textcolor{green}{\ding{52}} & \ding{56} & 6,387\\
MemeIntent \cite{park2024memeintent}  & explain: metaphors                                     & \textcolor{green}{\ding{52}} & \ding{56} & 950\\
\hdashline
\multirow{2}{*}{\textbf{MemeReaCon (ours)}} & \textbf{classify: meme + post + comment type/affection} & \multirow{2}{*}{\textcolor{green}{\ding{52}}} & \multirow{2}{*}{\textcolor{green}{\ding{52}}} & \multirow{2}{*}{\textbf{1,565}}\\
& \textbf{explain: meme + metaphors + post + post intents}\\
\bottomrule
\end{tabular}}
\caption{Comparisons with other related meme benchmarks.}
\label{tab:related_benchmarks}
\end{table*}

\section{Related Works}

\paragraph{Meme Classification.} 
The detection of harmful memes has emerged as a significant research area, supported by extensive benchmark datasets \cite{kiela2019supervised, pramanick2021detecting, lin2024goat} and community initiatives such as Facebook's Hateful Memes Challenge \cite{kiela2020hateful}. Research in this domain has evolved along several trajectories. Early approaches employed two-stream architectures that separately encode textual and visual features before applying attention mechanisms and multimodal fusion techniques for classification \cite{kiela2019supervised, suryawanshi2020multimodal, pramanick2021momenta}. A parallel line of work has focused on fine-tuning pre-trained multimodal models specifically for harmful content detection \cite{lippe2020multimodal, velioglu2020detecting, hee2022explaining, hee2023decoding}. Both methods are conducted on multiple harmful categories such as trolling \cite{suryawanshi2020multimodal}, hateful \cite{kiela2020hateful}, anti-semitism \cite{chandra2021subverting}, misogynous \cite{fersini2022semeval}, and anti-vaccinationism \cite{knuutila2024spread}.


\paragraph{Meme Explanation.} Another stream of research focuses on understanding memes as standalone units. Tasks include generating textual explanations \cite{sharma2023you} or captions for memes \cite{hwang2023memecap}, classifying their sentiment or evoked emotions \cite{hee2023decoding}, identifying depicted entities \cite{sharma2023you}, or explaining their underlying humor \cite{sharma2022findings}. These studies typically operate on decontextualized memes, removing them from the original posts and discussions where their meaning is shaped and negotiated. This methodological choice inherently limits the ability to assess if models grasp the social function of the meme (i.e., \textbf{why} it was used \textbf{there}).

\paragraph{MemeReaCon's Position.} Table \ref{tab:related_benchmarks} shows related meme benchmarks. MemeReaCon occupies a unique position by being the first benchmark, to our knowledge, specifically constructed to evaluate the fine-grained contextual reasoning required to understand memes as they are used in online posts. It mandates the integration of the meme image and the full original post text. Its detailed annotations concerning the context-meme relationship, meme structure, and comment interactions enable a more nuanced analysis of LVLM capabilities and failures than previously possible.

\section{Constructing the MemeReaCon Benchmark}
The central goal of MemeReaCon is to provide a robust resource for evaluating the contextual reasoning capabilities of LVLMs when interpreting memes. Achieving this requires a dataset that is not only large and diverse but also curated about the interplay between a meme and its surrounding textual context. The construction process is detailed below.

\begin{figure*}
    \centering
    \includegraphics[width=\linewidth]{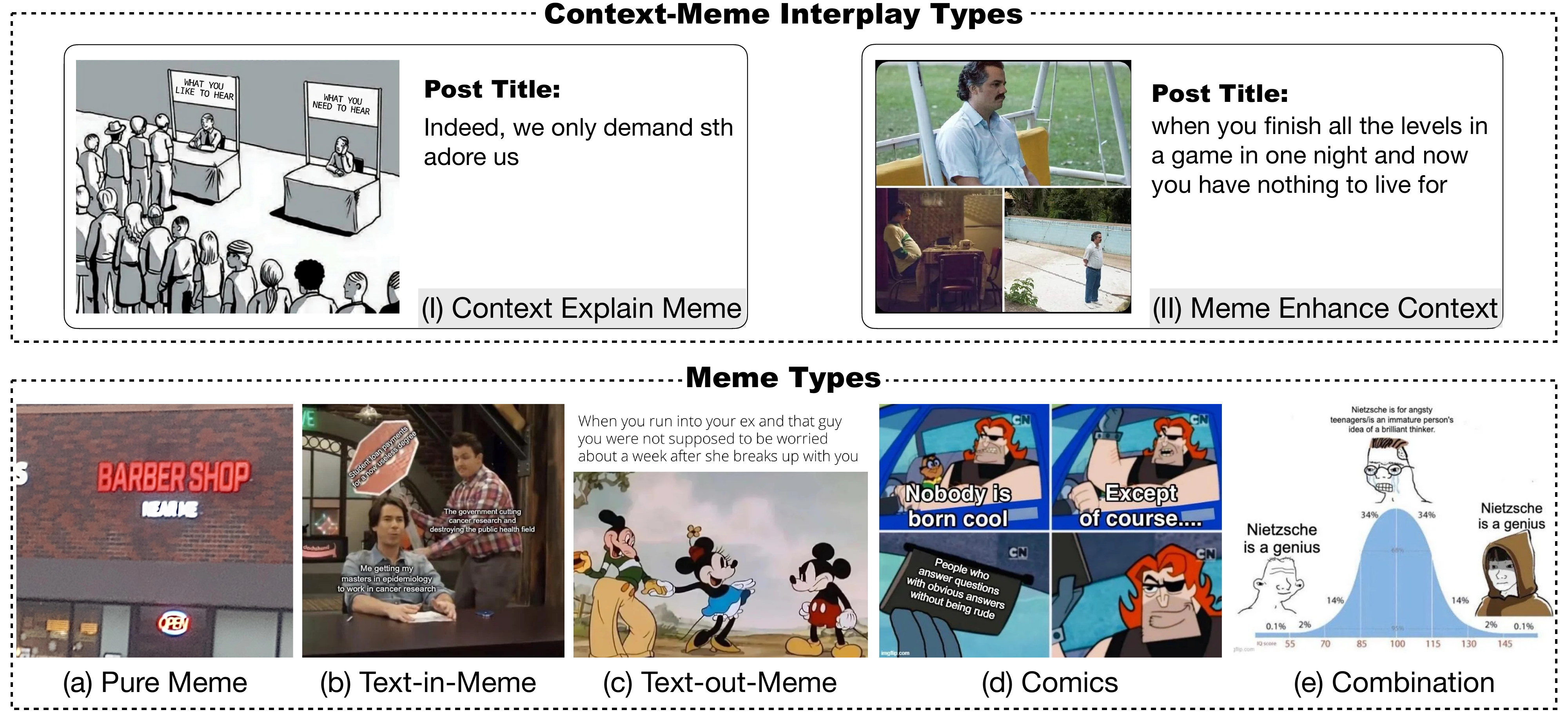}
    \caption{Cases of each annotation scheme. Top-side (I) and (II) represent the label of Context-Meme Interplay (CMI). Bottom-side (a) to (e) show the label of Meme Composition (MC).}
    \label{fig:demo_show_data}
\end{figure*}

\subsection{Data Collection}

To capture authentic meme usage patterns within varied contexts, we selected Reddit\footnote{\url{https://www.reddit.com}} as our primary data source. Reddit hosts a vast number of communities with distinct topics and communication styles, making it an ideal ecosystem for observing how the same meme template might be interpreted differently across contexts. We specifically chose five diverse, high-activity, English subreddits to ensure broad coverage:
\begin{itemize}[leftmargin=*, itemsep=0pt]
    \item \textbf{r/memes} and \textbf{r/meme}: Two large, general-purpose communities offering a baseline of popular meme formats and topics.
    \item \textbf{r/ProgrammerHumor}: A niche community focused on technology and programmer-specific context and humor.
    \item \textbf{r/BritishMemes}: A culturally specific community, requiring understanding of UK-related references, stereotypes, and events.
    \item \textbf{r/RelationshipMemes}: A social community centered on dating and interpersonal dynamics, often involving nuanced emotional expression.
\end{itemize}
This curated selection ensures variability in the types of contextual information (general knowledge, technical terms, cultural references, and social cues) required for successful interpretation.

We collected publicly available posts submitted between January 2022 and May 2025 using the Python Reddit API Wrapper. Our initial query targeted posts containing: (i) a textual title, (ii) an associated meme image, and (iii) the top-rated comments to filter out posts with community interaction. This initial pool contained over 3,000 potential candidates.

\subsection{Filtering for Quality and Contextual Relevance}
The raw data required careful filtering to isolate instances suitable for evaluating contextual reasoning. Our multi-stage filtering process aimed to maximize data quality and ensure that each instance contained sufficient context for meaningful analysis.

Firstly, we removed posts that were deleted (by user or admin), associated with suspended accounts, or contained broken image links. This step ensured the integrity and reproducibility of the dataset instances. Approximately 24\% of the initial pool was removed here.

Besides, to ensure presence of textual context accompanying the meme. We filtered out posts with very short context (fewer than 3 words\footnote{Some of contexts were internet-cultural abbreviations containing less than 3 words. We include these strong-cultural abbreviations too.}), as these often lack the necessary linguistic cues to establish a specific context beyond the meme image itself. This step removed roughly 18\% of the remaining posts, focusing the dataset on instances where textual context is explicitly provided.

While sourcing from meme-centric subreddits increases the likelihood of collecting actual memes, we implemented a verification step during annotation. Annotators removed non-meme images (e.g., selfie, advertisements) (in approximately 8\% of filtered posts).

Then, for comments, we selected the single highest-voted, non-deleted comment (excluding bot comments) as a proxy for the dominant community reaction or interpretation. To ensure the comment provided substantive feedback, we required a minimum length of 3 words. Posts lacking such a comment were also included noted as [none].

Each resulting instance was structured to include the meme image, the post title, the post body (marked empty if absent), and the selected top comment text. All usernames were anonymized to protect user privacy.

\begin{figure*}[!t]
    \centering
    \includegraphics[width=.75\linewidth]{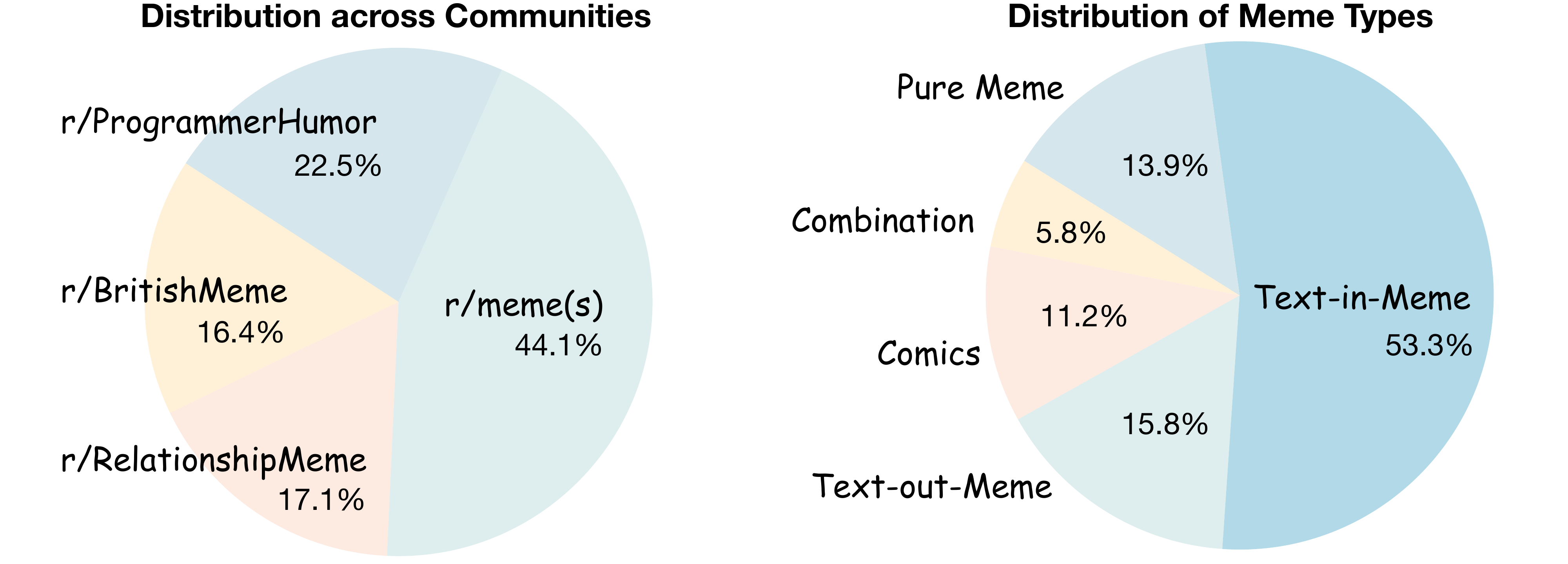}
    \caption{Statistics of our MemeReaCon. Our MemeReaCon benchmark comprises 1,565 annotated instances collected from five diverse subreddits. Detailed statistics can be found in Appendix~\ref{apd:statistics_of_memere}.}
    \label{fig:meme_stat}
\end{figure*}

\begin{figure}[!t]
    \centering
    \includegraphics[width=.9\linewidth]{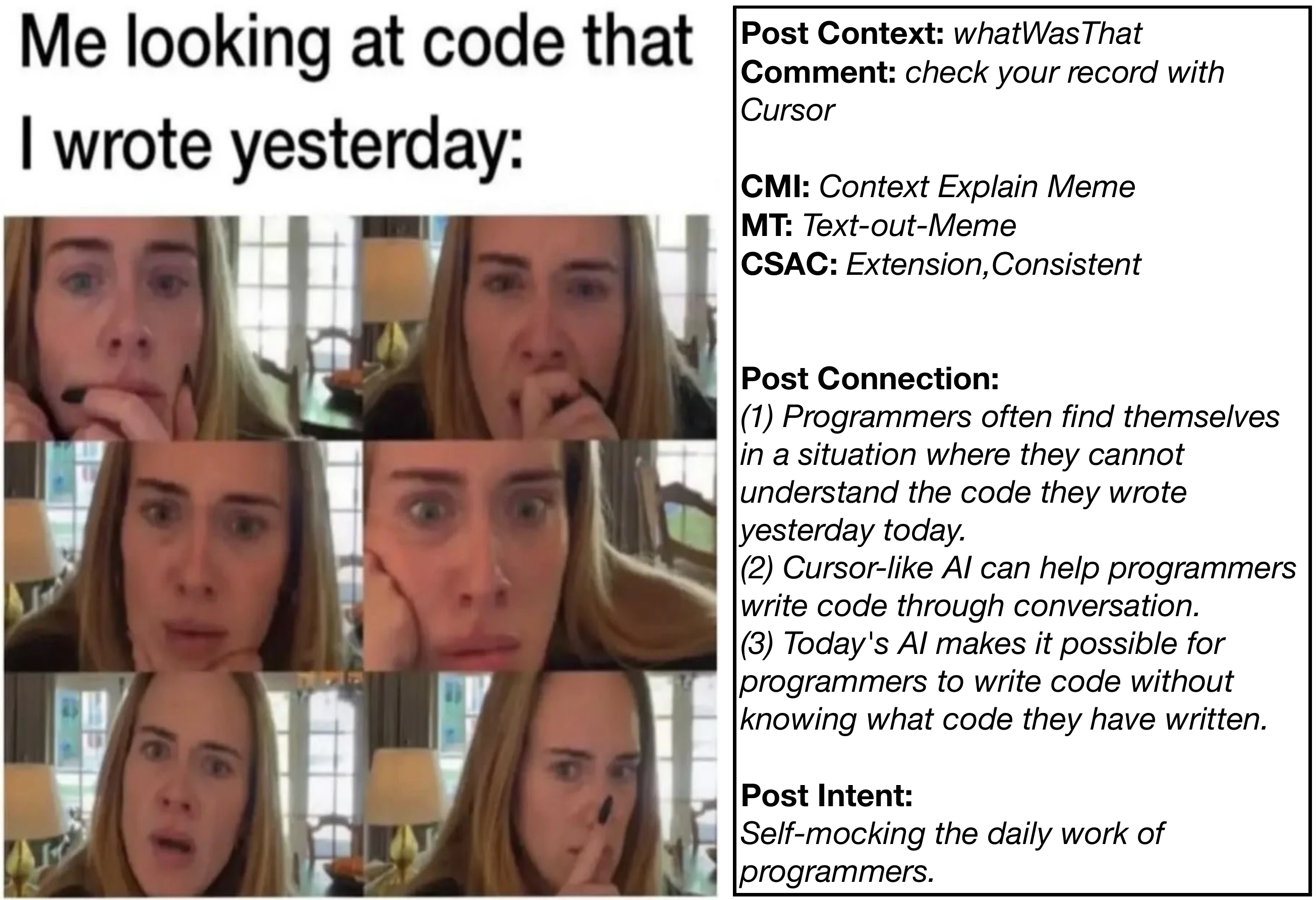}
    \caption{Example of a meme in MemeReaCon.}
    \label{fig:demo_show_all_annotation_scheme}
\end{figure}

\subsection{Annotation Scheme}

The annotation scheme is designed specifically to target the reasoning processes involved in understanding a meme within its post context. We developed labels that move beyond simple classification to capture the nuances of the context-meme connection and its intent. Our scheme includes five key dimensions:
\begin{itemize}[leftmargin=*, itemsep=0pt]
    \item \textbf{Context-Meme Interplay (CMI):} to directly addresses the question: how does the context relate to the meme (shown in Figure~\ref{fig:demo_show_data} (I) and (II))?
        \begin{itemize}[leftmargin=*, itemsep=0pt]
            \item \textit{Context Explain Meme (CEM)}: The text is essential for understanding the meme's relevance or specific meaning.
            \item \textit{Meme Enhance Context (MEC)}: The text establishes a point, and the meme serves to illustrate, emphasize, or add humor/emotion.
        \end{itemize}
    \item \textbf{Meme Types (MT):} to understand how information is distributed in meme (shown in Figure~\ref{fig:demo_show_data} (a) to (e)).
        \begin{itemize}[leftmargin=*, itemsep=0pt]
            \item \textit{Pure Meme}: Visuals carry the primary load.
            \item \textit{Text-in-Meme}: Embedded text is integral.
            \item \textit{Text-out-Meme}: Post title/body acts as the primary caption for a reusable template.
            \item \textit{Comics}: Multi-panel narrative structure.
            \item \textit{Combination}: Multi-type figures are combined together to perform the unitary meaning.
        \end{itemize}
    \item \textbf{Comment Stance and Affective Consistence (CSAC):} stance to assess the relationship between the top comment and the post. Affective consistence to assess the affection of a comment between its literal and its intended meaning.
        \begin{itemize}[leftmargin=*, itemsep=0pt]
            \item [] (1) From stance-level:
            \item \textit{Support}: Agrees with or reinforces the post.
            \item \textit{Deny}: Disagrees with or challenges the post.
            \item \textit{Extension}: Builds upon the post.
            \item [] (2) From affection-level:
            \item \textit{Consistent}: Same to its intent affection.
            \item \textit{Inconsistent}: Different from its literal one to perform a sarcastic or complain.
        \end{itemize}
    \item \textbf{Post Connection (PC):} to capture the logical or thematic linkages among the post context, meme, and comments, provided in key points that identify the specific connections between elements.
    \item \textbf{Post Intent (PI):} to identify the author's purpose for creating and sharing the post, such as humor, experience sharing, and complaint.
\end{itemize}

\paragraph{Annotation Process and Quality Control.} Figure~\ref{fig:demo_show_all_annotation_scheme} shows an example of our MemeReaCon\footnote{More cases are shown in Appendix~\ref{apd:more_cases}.}. Ensuring high-quality annotations was prominent. We recruited and trained 6 annotators (English-speaking Ph.D. students familiar with internet culture) using detailed guidelines and iteratively trained on 200 samples. The main annotation was conducted via a customized web interface displaying all components. To maximize reliability, each instance was independently annotated by 3 annotators. Disagreements were resolved by majority vote. For the rare cases of complete disagreement (3 unique labels for an instance), a senior annotator determined based on the guidelines and discussion. We calculated inter-annotator agreement (IAA) using Fleiss' Kappa ($\kappa$) on a held-out set of 500 instances annotated by all 6 annotators prior to the main task. The achieved agreement was substantial: CMI ($\kappa=0.86$), MT ($\kappa=0.88$), CSAC ($\kappa=0.75$), PC ($\kappa=0.79$), and PI ($\kappa=0.81$), indicating the robustness and clarity of our annotation scheme and process.

\subsection{Dataset Statistics}
The final MemeReaCon benchmark comprises 1,565 annotated instances collected from five diverse subreddits. Figure~\ref{fig:meme_stat} provides statistics of our MemeReaCon. Detailed statistics can be found in Appendix \ref{apd:statistics_of_memere}.

\section{Experiments}
\label{sec:experiments}

Our experiments with MemeReaCon are designed to address two key research questions: (1) to what extent do models understand the meme? (2) to what extent does the post affect models' understanding of meme?

\subsection{Experimental Setup}

\paragraph{Models Evaluated.}
We evaluated 10 diverse state-of-the-art models spanning three architectural paradigms, alongside two unimodal baselines to establish comparative foundations:

\begin{itemize}[leftmargin=*, itemsep=0pt]
    \item \textbf{Unimodal Baselines:} Qwen2.5 \cite{yang2024qwen2} (text-only) and Flamingo \cite{alayrac2022flamingo} (image-only) establish performance boundaries for single-modality reasoning.
    \item \textbf{Vision-Language Models (VLM):} LLaVA-OneVision-7B \cite{li2024llava}, Phi-4-MM-5.6B \cite{abdin2024phi}, Qwen2.5-VL-7B \cite{bai2025qwen2}, Qwen2.5-Omni-7B \cite{xu2025qwen2}, and InternVL3-8B \cite{chen2024internvl} represent approaches where vision and language capabilities are jointly trained.
    \item \textbf{Vision Reasoning Models (VRM):} QvQ-72B \cite{qvq}, GPT-4o \cite{hurst2024gpt}, Grok3 \cite{grok3}, Claude-3.7-sonnet-thinking \cite{claude37}, and Gemini-2.5-Pro \cite{gemini25pro} integrate advanced reasoning mechanisms atop vision-language foundations, representing the current frontier.
\end{itemize}

\begin{table*}[!t]
\centering
\adjustbox{max width=\linewidth}{
\begin{tabular}{lcccccccc}
\toprule
\multirow{2}{*}{Model} & \multicolumn{2}{c}{CMI-C} & \multicolumn{2}{c}{CSAC-C} & \multicolumn{2}{c}{PC-G} & \multicolumn{2}{c}{PI-G}\\ 
\cmidrule(lr){2-3} \cmidrule(lr){4-5} \cmidrule(lr){6-7} \cmidrule(lr){8-9}
 & Acc (\%) & MacF1 (\%) & Acc (\%) & MacF1 (\%) & B-S (\%) & R-L (\%) & B-S (\%) & R-L (\%)\\ 
\midrule
\multicolumn{9}{l}{\textit{Unimodal Baselines}} \\
Qwen2.5  & 54.83 & 53.92 & 59.27 & 41.24 & 46.48 & 46.37 & 22.63 & 17.82 \\
Flamingo & 52.14 & 51.58 & 31.73 & 22.79 & 25.42 & 18.13 & 9.31 & 8.47 \\
\midrule
\multicolumn{9}{l}{\textit{Vision-Language Models (VLM)}} \\
LLaVA-OneVision & 56.32 & 55.76 & 38.91 & 29.08 & 20.19 & 22.53 & 12.68 & 10.91 \\
Phi-4-MM        & 58.47 & 58.12 & 42.34 & 32.61 & 26.92 & 28.39 & 15.52 & 13.26 \\
InternVL3       & 64.72 & 64.18 & 49.53 & 38.41 & 37.23 & 38.92 & 25.63 & 20.43 \\
Qwen2.5-VL      & 61.43 & 60.97 & 46.15 & 35.83 & 33.74 & 31.27 & 19.76 & 16.04 \\
Qwen2.5-Omni    & 68.86 & 68.47 & 54.32 & 42.13 & 43.04 & 44.79 & 30.43 & 24.18 \\
Qwen2.5-Omni w/ CoT & 71.35 & 70.96 & 57.68 & 44.72 & 47.19 & 48.23 & 33.76 & 26.95 \\
Qwen2.5-Omni w/ SC & 73.42 & 73.09 & 60.17 & 46.89 & 49.87 & 50.42 & 36.28 & 29.37 \\
\midrule
\multicolumn{9}{l}{\textit{Vision Reasoning Models (VRM)}} \\
gpt-4o     & 72.48 & 71.96 & 58.76 & 46.53 & 57.82 & 48.47 & 34.94 & 28.37 \\
QvQ        & 75.29 & 74.87 & 62.61 & 49.74 & 60.13 & 51.25 & 39.52 & 32.64 \\
Grok-3     & 78.36 & 78.09 & 65.73 & 53.21 & 62.29 & 54.03 & 43.19 & 36.47 \\
Claude-3.7 & 80.97 & 80.56 & 68.39 & 55.93 & 64.48 & 57.14 & 47.59 & 40.31 \\
Gemini-2.5-pro & \textbf{83.21} & \textbf{82.86} & \textbf{71.28} & \textbf{59.42} & \textbf{66.89} & \textbf{60.38} & \textbf{52.34} & \textbf{44.86} \\
\bottomrule
\end{tabular}}
\caption{Performance comparison across model architectures on MemeReaCon tasks. \textbf{Bold} indicates the best performance.}
\label{tab:overall_results}
\end{table*}

\vspace{-1em}
\paragraph{Evaluation Settings.} All evaluations were conducted in a zero-shot setting with no fine-tuning. For classification tasks, we report accuracy and macro F1-score to account for class imbalance. For generative tasks, we use BERTScore (B-S) \cite{DBLP:conf/iclr/ZhangKWWA20} and ROUGE-L (R-L) to evaluate semantic and lexical similarity.

\paragraph{Tasks.}
We designed four primary tasks of increasing complexity to systematically probe different dimensions of contextual meme understanding.

It is important to note the role of the \textbf{Meme Types (MT)} annotation. While MT is a crucial dimension for understanding the structural properties of memes, we do not define a direct classification task for it. Instead, MT serves as an analytical lens through which we evaluate model performance on the other defined tasks. This allows for a fine-grained analysis of how different meme structures impact a model's ability.

The four primary evaluation tasks are:

\begin{itemize}[leftmargin=*, itemsep=0pt]
    \item \textbf{Context-Meme Interplay Classification (CMI-C):} Given the post context and the meme, models must classify the relationship as either \textit{Context Explain Meme (CEM)} or \textit{Meme Enhance Context (MEC)}. This task evaluates the model's basic understanding of how textual context and visual meme content depend on each other.
    \item \textbf{Comment Stance and Affective Consistent Classification (CSAC-C):} This is a two-part classification task. Given the original post (context + meme) and a top-level comment, models must: (1) determine the comment's stance towards the post (\textit{Support}, \textit{Deny}, or \textit{Extension}), and (2) identify whether the comment's literal affection is \textit{Consistent} or \textit{Inconsistent} with its intended meaning. This task probes deeper social reasoning capabilities, including the ability to understand agreement, disagreement, and nuanced expressions like sarcasm.
    \item \textbf{Post Connection Generation (PC-G):} Given the post context, the meme, and a set of relevant comments, models are required to generate a free-form text. This text should explain the key logical or thematic connections linking these elements. This generative task evaluates the model's overall understanding and its ability to articulate the reasoning chain.
    \item \textbf{Post Intent Generation (PI-G):} Based on all available evidence (post context, meme, and comments), models must generate the original poster's communicative intent (e.g., humor, complaint). This task assesses the model's ability to understand the overall purpose of the multimodal post.
\end{itemize}

These tasks are designed to progressively challenge models, moving from classifying direct relationships (CMI-C) to understanding complex social cues (CSAC-C), generating coherent explanations (PC-G), and inferring high-level intent (PI-G). Together, they provide a comprehensive benchmark for evaluating contextual reasoning abilities in the domain of internet memes. Detailed implementations can be found in Appendix~\ref{apd:implementations}.

\subsection{Overall Performance Comparison}
\label{sec:overall_performance}

Table~\ref{tab:overall_results} presents a comprehensive performance evaluation of various models on our MemeReaCon benchmark. Our analysis reveals critical insights into current LVLM capabilities and limitations in understanding social media posts with memes.

\begin{figure*}[!t]
\centering
\includegraphics[width=\linewidth]{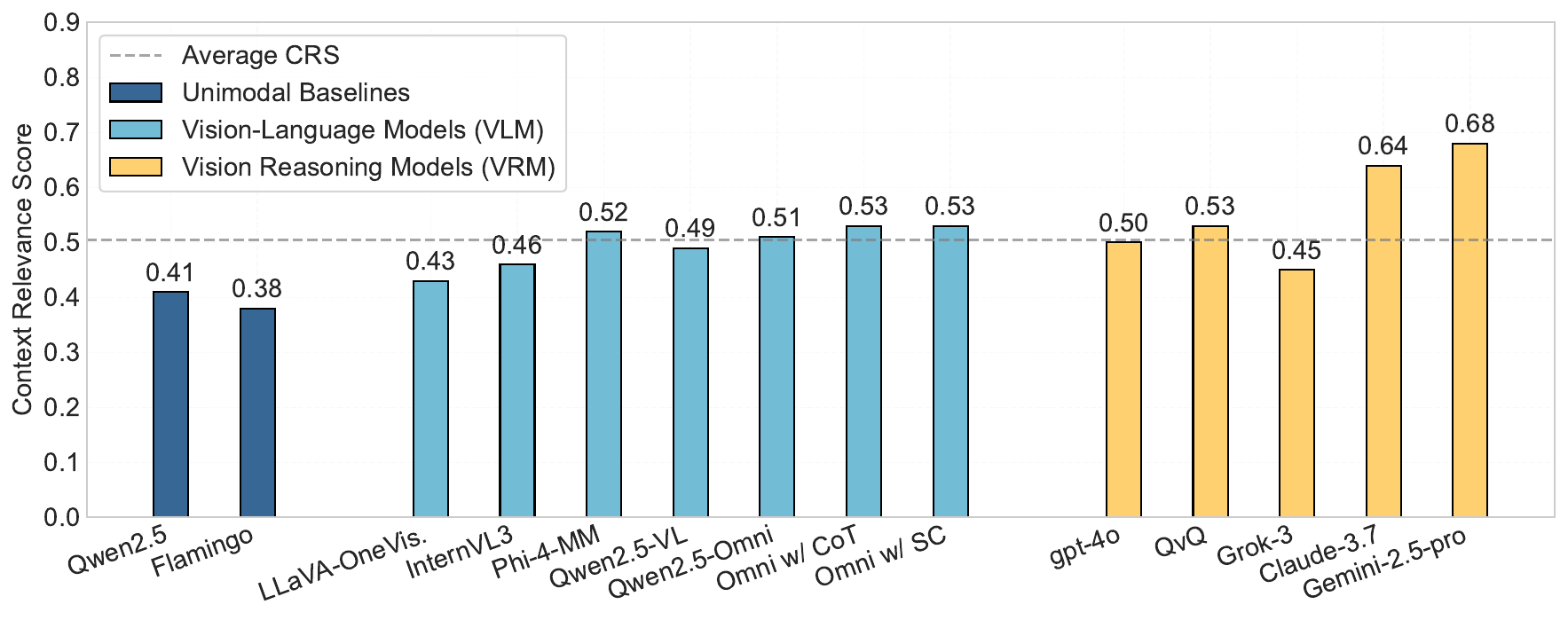}
\caption{Context relevance scores across model categories, measuring how effectively models integrate information from multiple contextual sources.}
\label{fig:context_integration}
\end{figure*}

\paragraph{Surface-level Understanding vs. Deep Comprehension.} 
While models demonstrate reasonable proficiency on simpler classification tasks (CMI-C, CSAC-C), their performance deteriorates substantially on generative tasks requiring deeper post comprehension (PC-G, PI-G). Even the top-performing Gemini-2.5-pro shows a big drop from classification (83.21\% accuracy on CMI-C) to generative tasks (60.38\% ROUGE-L on PC-G, 44.86\% on PI-G). This performance cliff indicates that current models can identify superficial relationships between text and images but struggle to synthesize holistic interpretations that capture the post's communicative intent and social context. The low PI-G scores particularly suggest that current models still fall short in understanding the nuanced social dynamics embedded in meme-based communication.

When applying Chain-of-Thought (CoT) and Self-Consistency (SC) techniques to Qwen2.5-Omni, we observe modest improvements across all tasks. However, these enhancements are more for classification tasks (+4.56\% on CMI-C with SC) and less impactful for generative tasks (+3.85\% on BERTScore on PI-G). This suggests that while structured reasoning approaches can help models better classify relationships, they offer limited benefits for the deeper contextual integration needed to understand post meaning and intent.

\begin{figure*}[!t]
\centering
\includegraphics[width=.95\linewidth]{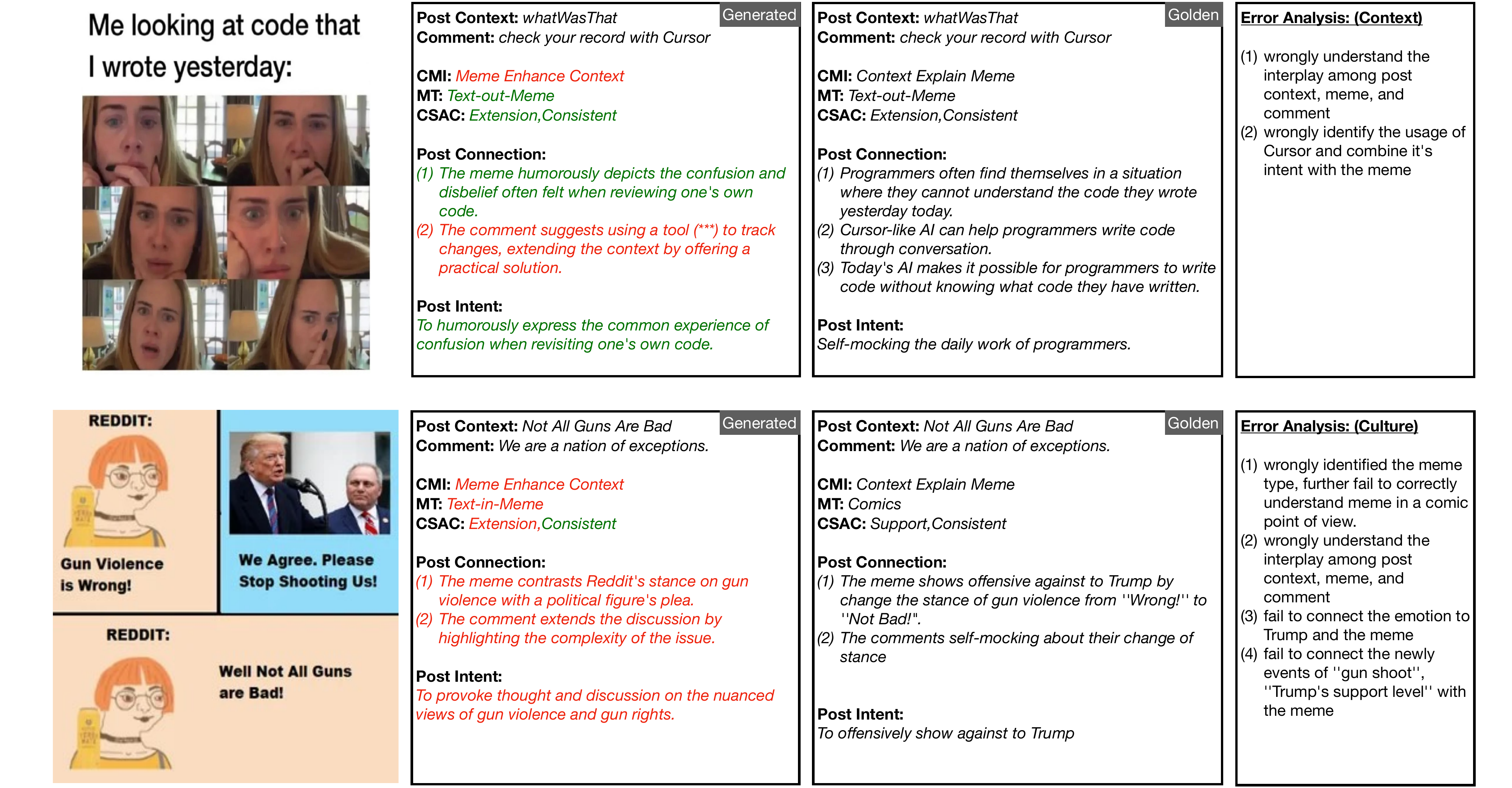}
\caption{Illustration of some cases in error. The green text indicates the correct answer. The red text indicates the wrong answer.}
\label{fig:error_cases}
\end{figure*}

\vspace{-1em}
\paragraph{Post Components Integration Challenge.} 
To quantitatively assess models' ability to integrate information across modalities and contextual elements, we introduce the Context Relevance Score (CRS), defined as:
\begin{equation}
\text{CRS} = \frac{1}{N}\sum_{i=1}^{N} w_i \cdot \text{Rel}(r_i, \{c_j\}_{j=1}^M),
\end{equation}
where $N$ is the number of evaluation samples, $r_i$ is the model's response for sample $i$, $\{c_j\}_{j=1}^M$ represents the $M$ contextual elements (post text, image, comments) for sample $i$, $\text{Rel}(\cdot)$ measures the semantic relevance between the response and all contextual elements (computed using BERTScore with a threshold of 0.7 for relevance), and $w_i$ is a difficulty weight based on the number of contextual elements requiring integration. CRS ranges from 0 to 1, with higher scores indicating better cross-contextual integration.

Our CRS analysis reveals significant gaps in contextual integration capabilities. As shown in Figure~\ref{fig:context_integration}, VRMs achieve higher CRS values compared to VLMs. But the best models struggle with fully integrating information across modalities and contextual elements. This finding aligns with the poor performance on PC-G and PI-G tasks, confirming that contextual integration represents a fundamental bottleneck in current architectures. We show more analysis of performance in different communities (\ref{apd:community_specific_performance_analysis}), meme structure (\ref{apd:meme_structure_analysis}), meme text-density (\ref{apd:meme_text-density_analysis}), comment affection (\ref{apd:comment_affection_analysis}), and modality contribution (\ref{apd:modality_analysis}).

\subsection{Error Analysis}
\label{sec:errors_analysis}

To gain deeper insights into how post context influences meme interpretation, we conducted a systematic error analysis across all evaluated models. This analysis reveals critical limitations in current models when processing contextually embedded memes and highlights failure patterns that occur at the intersection of visual humor and social context.

We categorized errors into four distinct patterns that emerged consistently across models: context error, visual error, semantic error, and cultural error. Appendix~\ref{apd:error} shows detailed definition of these patterns and distributions of error types across models. Figure~\ref{fig:error_cases} shows the selected error cases. More cases can be found in Appendix~\ref{apd:more_error_cases}.

\section{Conclusion}
In this paper, we introduced MemeReaCon, a novel benchmark that addresses a critical gap in meme understanding research by preserving the post context for meme interpretation. Our findings revealed significant limitations in current LVLMs to integrate contextual information when explaining memes, with models often failing to establish meaningful connections between visual content and surrounding context or overlooking communicative purpose in favor of surface-level visual analysis. Besides, by identifying the dual relationship patterns between memes and their contexts, we provided a framework for evaluating how well models understand the diverse communicative functions of memes in online environments. This work not only highlights the context-insensitive limitations of current models but also establishes a foundation for future  to more accurately capture how humans naturally process and interpret memes within their original discourse contexts.

\section*{Limitations}

Our work, while comprehensive, is subject to certain limitations, primarily concerning the nuances of annotation when dealing with complex connections and intents and the inherent subjectivity in meme interpretation. First, regarding the annotation of post connections, we observed that the explicit post connections was less consistent across annotations in some cases. This suggests a challenge in achieving widespread mutual agreement on a precise methodology for connecting posters' context meaning with the meme meanings. Even when annotators possess the general knowledge to understand the meme's overall message, a shared, systematic approach to deconstructing and codifying the specific metaphorical knowledge embedded in the memes may not be uniformly applied. Second, the interpretation of memes is deeply depend on annotator's background knowledge, encompassing cultural, social, and contextual understanding, which inherently varies among annotators.

\section*{Ethics Statement}

The development of this benchmark for contextual meme understanding was guided by a commitment to responsible research practices. We have taken several steps to address potential ethical considerations related to data collection, annotation, and the potential impact of our work.

\paragraph{Data Collection and Provenance.}
The data for this benchmark was collected from Reddit, a publicly accessible platform, using its official Application Programming Interface (API). Our data collection adhered to Reddit's API terms of service. We focused on collecting posts that included both textual context and a meme image. To protect the privacy of Reddit users, all usernames and any other personally identifiable information (PII) were removed from the collected data. The dataset primarily consists of content that users have chosen to share publicly. We acknowledge that internet memes can sometimes contain sensitive or controversial themes.

\paragraph{Annotation Process and Annotator Considerations.}
The annotation of the collected data was performed by 6 Ph.D. students, all of whom are proficient English speakers and have a good understanding of internet culture and memes. Annotators were recruited from our research institution. Prior to commencing the annotation task, all annotators were provided with detailed guidelines and training on the annotation scheme to ensure consistency and quality. They were made aware of the research objectives and how their contributions would be used.

Recognizing that prolonged exposure to online content can sometimes be taxing, and that memes can vary widely in their subject matter, annotators were instructed that they could skip any specific data instance they felt uncomfortable annotating, without any penalty. The annotation tasks were designed to be objective, focusing on the relationship between context, meme, and comments. The PhD students involved in annotation were part of the broader research effort and their contribution is acknowledged; this work formed part of their research activities.

We paid \$0.19 for each data annotation. The annotators were compensated with an average hourly wage of \$14.82, which is comparable to the local minimum wage. We did not collect any personal information from annotators without their permission.

\bibliography{custom}

\begin{thebibliography}{34}
\providecommand{\natexlab}[1]{#1}

\bibitem[{Abdin et~al.(2024)Abdin, Aneja, Behl, Bubeck, Eldan, Gunasekar,
  Harrison, Hewett, Javaheripi, Kauffmann, and et~al.}]{abdin2024phi}
Marah Abdin, Jyoti Aneja, Harkirat Behl, S{\'e}bastien Bubeck, Ronen Eldan,
  Suriya Gunasekar, Michael Harrison, Russell~J Hewett, Mojan Javaheripi, Piero
  Kauffmann, and et~al. 2024.
\newblock Phi-4 technical report.
\newblock \emph{arXiv preprint arXiv:2412.08905}.

\bibitem[{Alayrac et~al.(2022)Alayrac, Donahue, Luc, Miech, Barr, Hasson, Lenc,
  Mensch, Millican, Reynolds, and et~al.}]{alayrac2022flamingo}
Jean-Baptiste Alayrac, Jeff Donahue, Pauline Luc, Antoine Miech, Iain Barr,
  Yana Hasson, Karel Lenc, Arthur Mensch, Katherine Millican, Malcolm Reynolds,
  and et~al. 2022.
\newblock Flamingo: a visual language model for few-shot learning.
\newblock \emph{Advances in neural information processing systems},
  35:23716--23736.

\bibitem[{Anthropic(2025)}]{claude37}
Anthropic. 2025.
\newblock Claude 3.7 sonnet and claude code.
\newblock \emph{https://www.anthropic.com/news/claude-3-7-sonnet}.

\bibitem[{Bai et~al.(2025)Bai, Chen, Liu, Wang, Ge, Song, Dang, Wang, Wang,
  Tang, and et~al.}]{bai2025qwen2}
Shuai Bai, Keqin Chen, Xuejing Liu, Jialin Wang, Wenbin Ge, Sibo Song, Kai
  Dang, Peng Wang, Shijie Wang, Jun Tang, and et~al. 2025.
\newblock Qwen2. 5-vl technical report.
\newblock \emph{arXiv preprint arXiv:2502.13923}.

\bibitem[{Chandra et~al.(2021)Chandra, Pailla, Bhatia, Sanchawala, Gupta,
  Shrivastava, and Kumaraguru}]{chandra2021subverting}
Mohit Chandra, Dheeraj Pailla, Himanshu Bhatia, Aadilmehdi Sanchawala, Manish
  Gupta, Manish Shrivastava, and Ponnurangam Kumaraguru. 2021.
\newblock “subverting the jewtocracy”: Online antisemitism detection using
  multimodal deep learning.
\newblock In \emph{Proceedings of the 13th ACM Web Science Conference 2021},
  pages 148--157.

\bibitem[{Chen et~al.(2024)Chen, Wu, Wang, Su, Chen, Xing, Zhong, Zhang, Zhu,
  Lu, and et~al.}]{chen2024internvl}
Zhe Chen, Jiannan Wu, Wenhai Wang, Weijie Su, Guo Chen, Sen Xing, Muyan Zhong,
  Qinglong Zhang, Xizhou Zhu, Lewei Lu, and et~al. 2024.
\newblock Internvl: Scaling up vision foundation models and aligning for
  generic visual-linguistic tasks.
\newblock In \emph{Proceedings of the IEEE/CVF conference on computer vision
  and pattern recognition}, pages 24185--24198.

\bibitem[{DeepMind(2025)}]{gemini25pro}
Google DeepMind. 2025.
\newblock Gemini 2.5 pro: Best for coding and complex prompts.
\newblock \emph{https://deepmind.google/technologies/gemini/pro/}.

\bibitem[{Fersini et~al.(2022)Fersini, Gasparini, Rizzi, Saibene, Chulvi,
  Rosso, Lees, and Sorensen}]{fersini2022semeval}
Elisabetta Fersini, Francesca Gasparini, Giulia Rizzi, Aurora Saibene, Berta
  Chulvi, Paolo Rosso, Alyssa Lees, and Jeffrey Sorensen. 2022.
\newblock Semeval-2022 task 5: Multimedia automatic misogyny identification.
\newblock In \emph{Proceedings of the 16th International Workshop on Semantic
  Evaluation (SemEval-2022)}, pages 533--549.

\bibitem[{Hee et~al.(2023)Hee, Chong, and Lee}]{hee2023decoding}
Ming~Shan Hee, Wen-Haw Chong, and Roy Ka-Wei Lee. 2023.
\newblock Decoding the underlying meaning of multimodal hateful memes.
\newblock In \emph{Proceedings of the Thirty-Second International Joint
  Conference on Artificial Intelligence}, pages 5995--6003.

\bibitem[{Hee et~al.(2022)Hee, Lee, and Chong}]{hee2022explaining}
Ming~Shan Hee, Roy Ka-Wei Lee, and Wen-Haw Chong. 2022.
\newblock On explaining multimodal hateful meme detection models.
\newblock In \emph{Proceedings of the ACM Web Conference 2022}, pages
  3651--3655.

\bibitem[{Huang et~al.(2024)Huang, Lin, Ziyan, Luo, Chen, and
  Ma}]{huang2024towards}
Jianzhao Huang, Hongzhan Lin, Liu Ziyan, Ziyang Luo, Guang Chen, and Jing Ma.
  2024.
\newblock Towards low-resource harmful meme detection with lmm agents.
\newblock In \emph{Proceedings of the 2024 Conference on Empirical Methods in
  Natural Language Processing}, pages 2269--2293.

\bibitem[{Hurst et~al.(2024)Hurst, Lerer, Goucher, Perelman, Ramesh, Clark,
  Ostrow, Welihinda, Hayes, Radford, and et~al.}]{hurst2024gpt}
Aaron Hurst, Adam Lerer, Adam~P Goucher, Adam Perelman, Aditya Ramesh, Aidan
  Clark, AJ~Ostrow, Akila Welihinda, Alan Hayes, Alec Radford, and et~al. 2024.
\newblock Gpt-4o system card.
\newblock \emph{arXiv preprint arXiv:2410.21276}.

\bibitem[{Hwang and Shwartz(2023)}]{hwang2023memecap}
EunJeong Hwang and Vered Shwartz. 2023.
\newblock Memecap: A dataset for captioning and interpreting memes.
\newblock In \emph{Proceedings of the 2023 Conference on Empirical Methods in
  Natural Language Processing}, pages 1433--1445.

\bibitem[{Kiela et~al.(2019)Kiela, Bhooshan, Firooz, Perez, and
  Testuggine}]{kiela2019supervised}
Douwe Kiela, Suvrat Bhooshan, Hamed Firooz, Ethan Perez, and Davide Testuggine.
  2019.
\newblock Supervised multimodal bitransformers for classifying images and text.
\newblock \emph{arXiv preprint arXiv:1909.02950}.

\bibitem[{Kiela et~al.(2020)Kiela, Firooz, Mohan, Goswami, Singh, Ringshia, and
  Testuggine}]{kiela2020hateful}
Douwe Kiela, Hamed Firooz, Aravind Mohan, Vedanuj Goswami, Amanpreet Singh,
  Pratik Ringshia, and Davide Testuggine. 2020.
\newblock The hateful memes challenge: Detecting hate speech in multimodal
  memes.
\newblock \emph{Advances in neural information processing systems},
  33:2611--2624.

\bibitem[{Knuutila et~al.(2024)Knuutila, George, Bright, George, and
  Howard}]{knuutila2024spread}
Aleksi Knuutila, Anna George, Jonathan Bright, Anna George, and Philip Howard.
  2024.
\newblock The spread of anti-vaccination memes on facebook.
\newblock In \emph{Multidisciplinary International Symposium on Disinformation
  in Open Online Media}, pages 86--100. Springer.

\bibitem[{Li et~al.(2024)Li, Zhang, Guo, Zhang, Li, Zhang, Zhang, Zhang, Li,
  Liu, and et~al.}]{li2024llava}
Bo~Li, Yuanhan Zhang, Dong Guo, Renrui Zhang, Feng Li, Hao Zhang, Kaichen
  Zhang, Peiyuan Zhang, Yanwei Li, Ziwei Liu, and et~al. 2024.
\newblock Llava-onevision: Easy visual task transfer.
\newblock \emph{arXiv preprint arXiv:2408.03326}.

\bibitem[{Lin et~al.(2024)Lin, Luo, Wang, Yang, and Ma}]{lin2024goat}
Hongzhan Lin, Ziyang Luo, Bo~Wang, Ruichao Yang, and Jing Ma. 2024.
\newblock \href {https://arxiv.org/abs/2401.01523} {Goat-bench: Safety insights
  to large multimodal models through meme-based social abuse}.
\newblock \emph{Preprint}, arXiv:2401.01523.

\bibitem[{Lippe et~al.(2020)Lippe, Holla, Chandra, Rajamanickam, Antoniou,
  Shutova, and Yannakoudakis}]{lippe2020multimodal}
Phillip Lippe, Nithin Holla, Shantanu Chandra, Santhosh Rajamanickam, Georgios
  Antoniou, Ekaterina Shutova, and Helen Yannakoudakis. 2020.
\newblock A multimodal framework for the detection of hateful memes.
\newblock \emph{arXiv preprint arXiv:2012.12871}.

\bibitem[{Liu et~al.(2022)Liu, Geigle, Krebs, and Gurevych}]{liu2022figmemes}
Chen Liu, Gregor Geigle, Robin Krebs, and Iryna Gurevych. 2022.
\newblock Figmemes: A dataset for figurative language identification in
  politically-opinionated memes.
\newblock In \emph{Proceedings of the 2022 conference on empirical methods in
  natural language processing}, pages 7069--7086.

\bibitem[{Milner(2012)}]{milner2012world}
Ryan~M Milner. 2012.
\newblock The world made meme: Discourse and identity in participatory media.

\bibitem[{Park et~al.(2024)Park, Nguyen, Li, Shrestha, Vu, Wang, and
  Ng}]{park2024memeintent}
Jeongsik Park, Khoi~PN Nguyen, Terrence Li, Suyesh Shrestha, Megan~Kim Vu,
  Jerry~Yining Wang, and Vincent Ng. 2024.
\newblock Memeintent: Benchmarking intent description generation for memes.
\newblock In \emph{Proceedings of the 25th Annual Meeting of the Special
  Interest Group on Discourse and Dialogue}, pages 631--643.

\bibitem[{Pramanick et~al.(2021{\natexlab{a}})Pramanick, Dimitrov, Mukherjee,
  Sharma, Akhtar, Nakov, and Chakraborty}]{pramanick2021detecting}
Shraman Pramanick, Dimitar Dimitrov, Rituparna Mukherjee, Shivam Sharma,
  Md~Shad Akhtar, Preslav Nakov, and Tanmoy Chakraborty. 2021{\natexlab{a}}.
\newblock Detecting harmful memes and their targets.
\newblock In \emph{Findings of the Association for Computational Linguistics:
  ACL-IJCNLP 2021}, pages 2783--2796.

\bibitem[{Pramanick et~al.(2021{\natexlab{b}})Pramanick, Sharma, Dimitrov,
  Akhtar, Nakov, and Chakraborty}]{pramanick2021momenta}
Shraman Pramanick, Shivam Sharma, Dimitar Dimitrov, Md~Shad Akhtar, Preslav
  Nakov, and Tanmoy Chakraborty. 2021{\natexlab{b}}.
\newblock Momenta: A multimodal framework for detecting harmful memes and their
  targets.
\newblock In \emph{Findings of the Association for Computational Linguistics:
  EMNLP 2021}, pages 4439--4455.

\bibitem[{Qwen(2024)}]{qvq}
Qwen. 2024.
\newblock Qvq: To see the world with wisdom.
\newblock \emph{https://qwenlm.github.io/blog/qvq-72b-preview/}.

\bibitem[{Sharma et~al.(2023)Sharma, Agarwal, Suresh, Nakov, Akhtar, and
  Chakraborty}]{sharma2023you}
Shivam Sharma, Siddhant Agarwal, Tharun Suresh, Preslav Nakov, Md~Shad Akhtar,
  and Tanmoy Chakraborty. 2023.
\newblock What do you meme? generating explanations for visual semantic role
  labelling in memes.
\newblock In \emph{Proceedings of the AAAI Conference on Artificial
  Intelligence}, volume~37, pages 9763--9771.

\bibitem[{Sharma et~al.(2022)Sharma, Suresh, Kulkarni, Mathur, Nakov, Akhtar,
  and Chakraborty}]{sharma2022findings}
Shivam Sharma, Tharun Suresh, Atharva Kulkarni, Himanshi Mathur, Preslav Nakov,
  Md~Shad Akhtar, and Tanmoy Chakraborty. 2022.
\newblock Findings of the constraint 2022 shared task on detecting the hero,
  the villain, and the victim in memes.
\newblock In \emph{Proceedings of the Workshop on Combating Online Hostile
  Posts in Regional Languages during Emergency Situations}, pages 1--11.

\bibitem[{Suryawanshi et~al.(2020)Suryawanshi, Chakravarthi, Arcan, and
  Buitelaar}]{suryawanshi2020multimodal}
Shardul Suryawanshi, Bharathi~Raja Chakravarthi, Mihael Arcan, and Paul
  Buitelaar. 2020.
\newblock Multimodal meme dataset (multioff) for identifying offensive content
  in image and text.
\newblock In \emph{Proceedings of the second workshop on trolling, aggression
  and cyberbullying}, pages 32--41.

\bibitem[{Velioglu and Rose(2020)}]{velioglu2020detecting}
Riza Velioglu and Jewgeni Rose. 2020.
\newblock Detecting hate speech in memes using multimodal deep learning
  approaches: Prize-winning solution to hateful memes challenge.
\newblock \emph{arXiv preprint arXiv:2012.12975}.

\bibitem[{Wang et~al.(2024)Wang, Huang, Liang, Tu, Yang, and Xu}]{wang2024they}
Bingbing Wang, Shijue Huang, Bin Liang, Geng Tu, Min Yang, and Ruifeng Xu.
  2024.
\newblock What do they “meme”? a metaphor-aware multi-modal multi-task
  framework for fine-grained meme understanding.
\newblock \emph{Knowledge-Based Systems}, 294:111778.

\bibitem[{xAI(2025)}]{grok3}
xAI. 2025.
\newblock Grok 3: The age of reasoning agents.
\newblock \emph{https://x.ai}.

\bibitem[{Xu et~al.(2025)Xu, Guo, He, Hu, He, Bai, Chen, Wang, Fan, Dang, and
  et~al.}]{xu2025qwen2}
Jin Xu, Zhifang Guo, Jinzheng He, Hangrui Hu, Ting He, Shuai Bai, Keqin Chen,
  Jialin Wang, Yang Fan, Kai Dang, and et~al. 2025.
\newblock Qwen2. 5-omni technical report.
\newblock \emph{arXiv preprint arXiv:2503.20215}.

\bibitem[{Yang et~al.(2024)Yang, Yang, Hui, Zheng, Yu, Zhou, Li, Li, Liu,
  Huang, and et~al.}]{yang2024qwen2}
An~Yang, Baosong Yang, Binyuan Hui, Bo~Zheng, Bowen Yu, Chang Zhou, Chengpeng
  Li, Chengyuan Li, Dayiheng Liu, Fei Huang, and et~al. 2024.
\newblock Qwen2 technical report.
\newblock \emph{CoRR}.

\bibitem[{Zhang et~al.(2020)Zhang, Kishore, Wu, Weinberger, and
  Artzi}]{DBLP:conf/iclr/ZhangKWWA20}
Tianyi Zhang, Varsha Kishore, Felix Wu, Kilian~Q. Weinberger, and Yoav Artzi.
  2020.
\newblock \href {https://openreview.net/forum?id=SkeHuCVFDr} {Bertscore:
  Evaluating text generation with {BERT}}.
\newblock In \emph{8th International Conference on Learning Representations,
  {ICLR} 2020, Addis Ababa, Ethiopia, April 26-30, 2020}. OpenReview.net.

\end{thebibliography}

\appendix

\section{More Cases}
\label{apd:more_cases}

\begin{figure*}[!t]
    \centering
    \includegraphics[width=\linewidth]{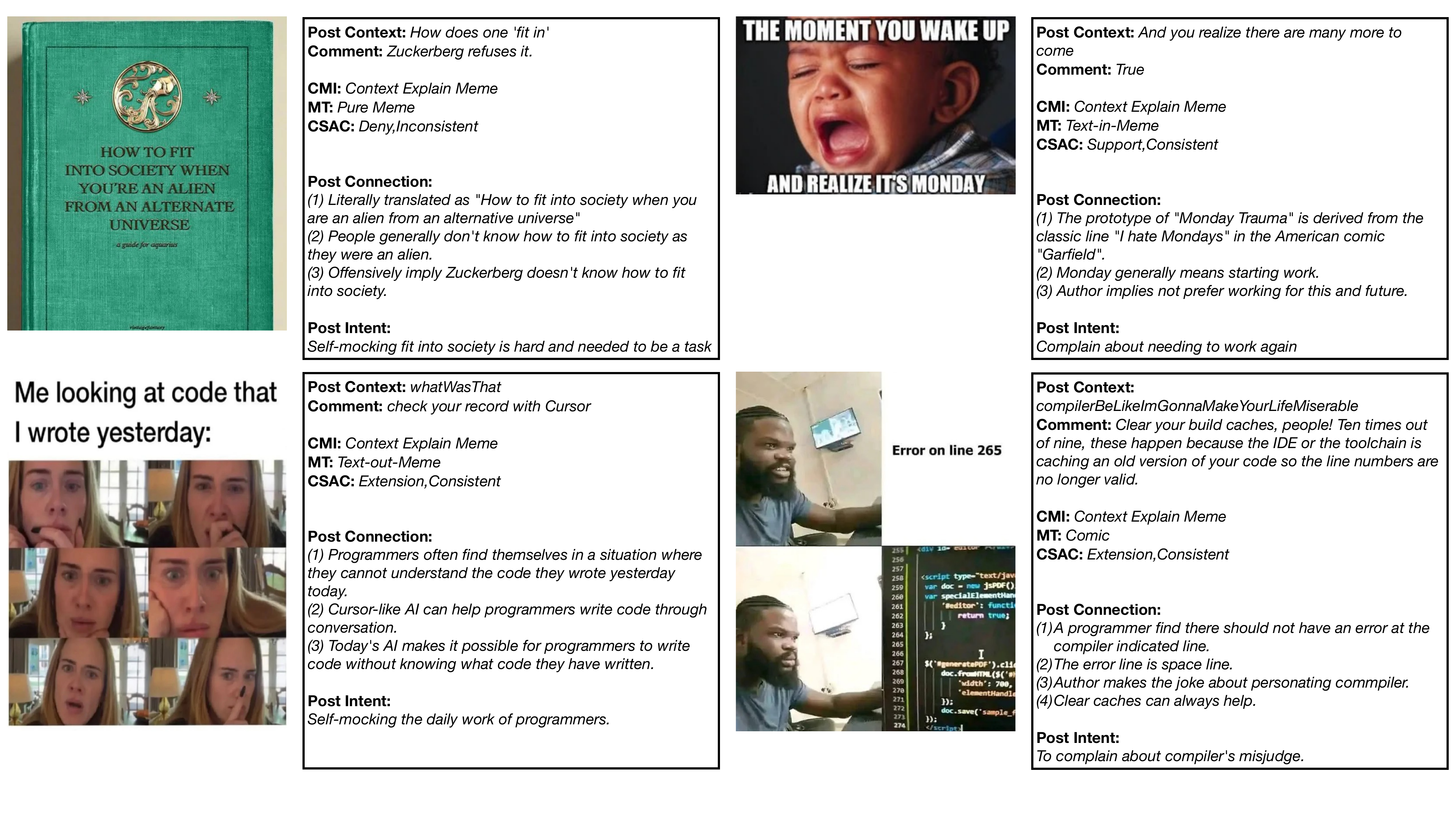}
    \caption{Examples of our proposed MemeReaCon (1/3).}
    \label{fig:more_case_1}
\end{figure*}

\begin{figure*}[!t]
    \centering
    \includegraphics[width=\linewidth]{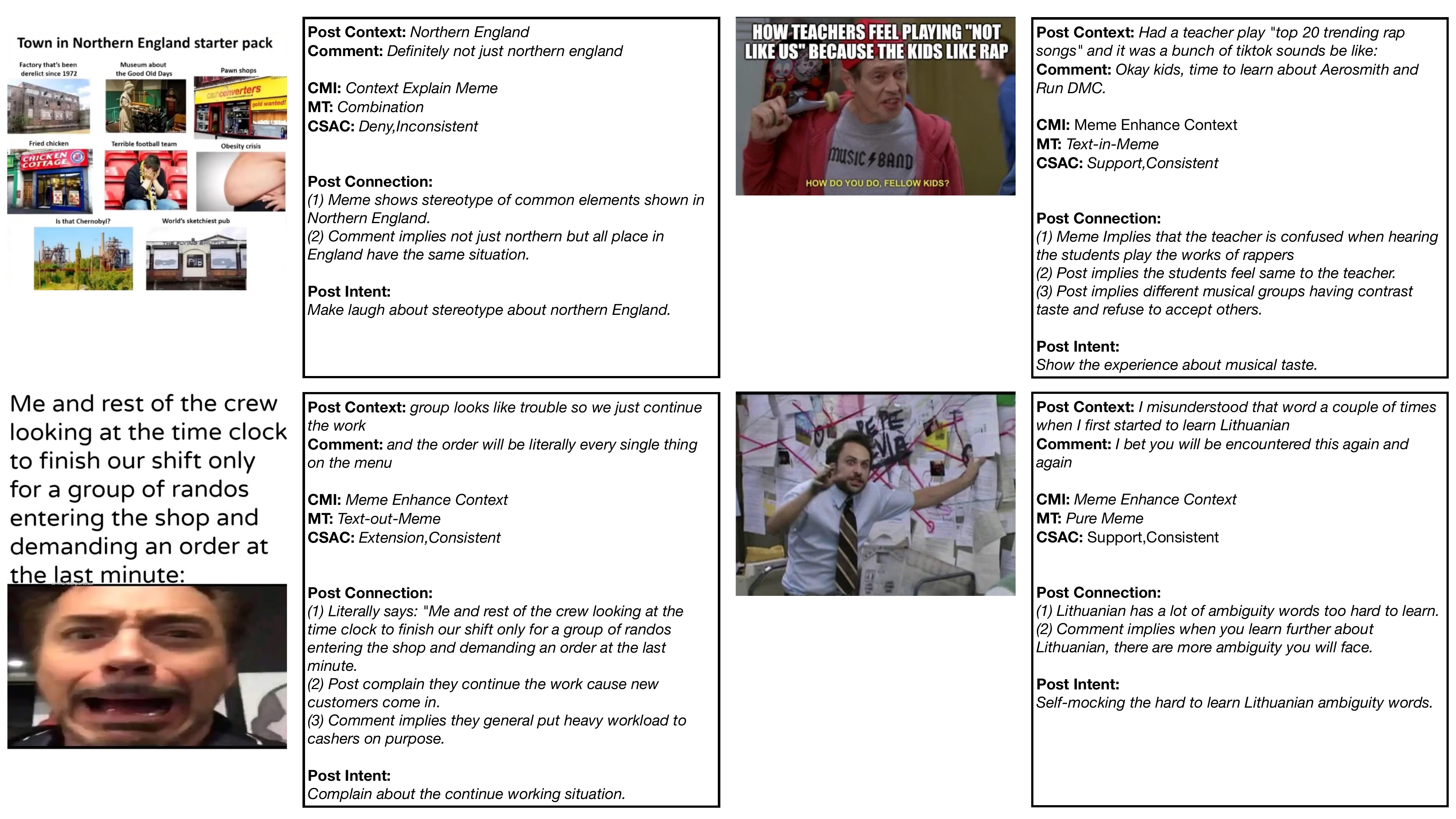}
    \caption{Examples of our proposed MemeReaCon (2/3).}
    \label{fig:more_case_2}
\end{figure*}

\begin{figure*}[!t]
    \centering
    \includegraphics[width=\linewidth]{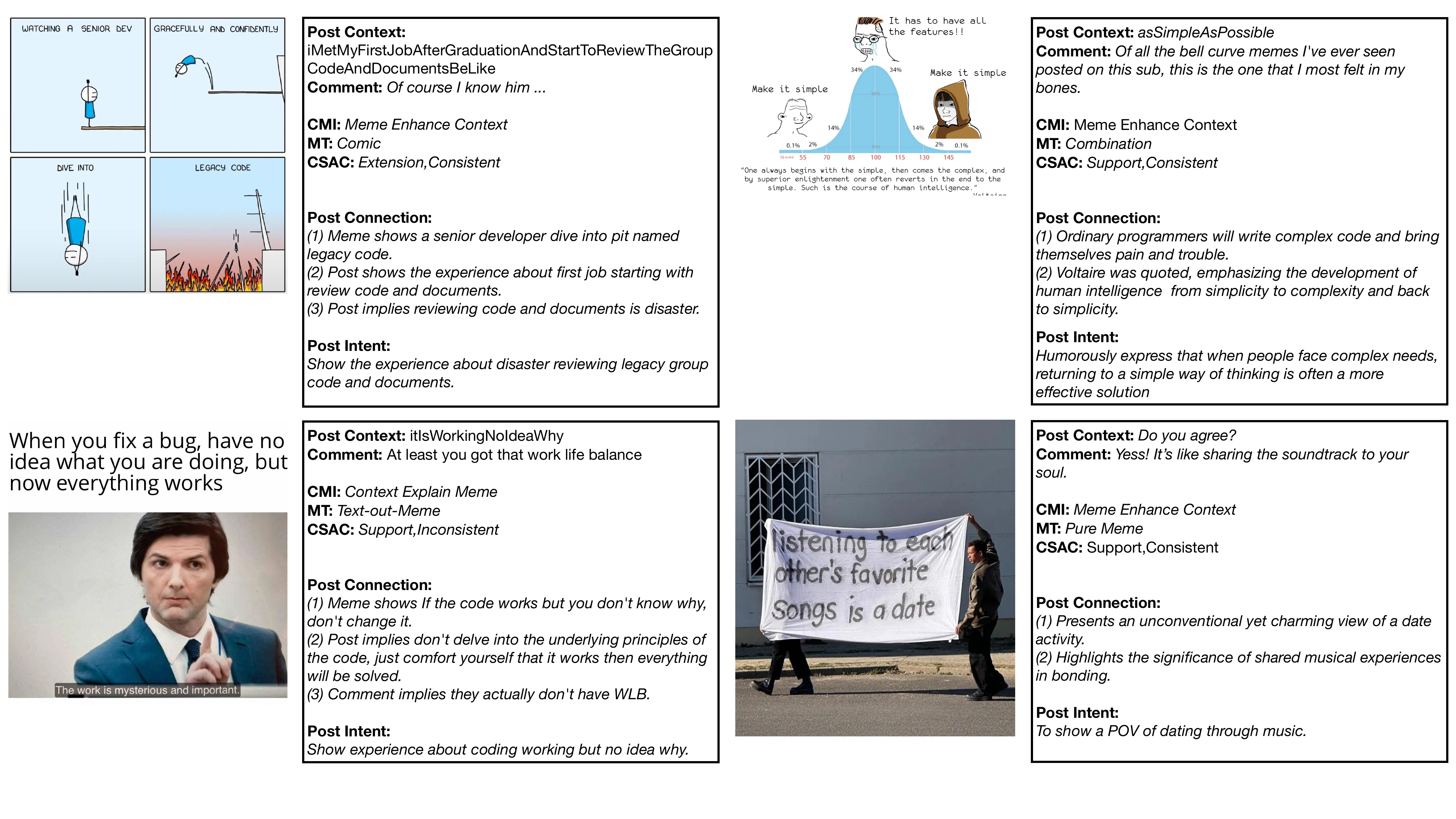}
    \caption{Examples of our proposed MemeReaCon (3/3).}
    \label{fig:more_case_3}
\end{figure*}

Figure~\ref{fig:more_case_1},~\ref{fig:more_case_2}, and~\ref{fig:more_case_3} shows more examples from our proposed MemeReaCon.

\section{Statistics of MemeReaCon}
\label{apd:statistics_of_memere}

\begin{table*}[!t]
\centering
\adjustbox{max width=\linewidth}{
\begin{tabular}{lrrrrr}
\toprule
\multirow{2}{*}{Category} & \multirow{2}{*}{Total (\%)} & \multicolumn{4}{c}{Distribution Across Subreddits}\\
\cmidrule(lr){3-6}
 &  & r/meme(s) & r/ProgrammerHumor & r/BritishMemes & r/RelationshipMemes\\
\midrule
\textbf{Overall Distribution} & \textbf{1565 (100\%)} & \textbf{690 (44.1\%)} & \textbf{352 (22.5\%)} & \textbf{256 (16.4\%)} & \textbf{267 (17.1\%)}\\
\midrule
\multicolumn{6}{l}{\textbf{Context-Meme Interplay (CMI)}} \\
\; Context Explains Meme & 796 (50.9\%) & 339 (49.1\%) & 187 (53.1\%) & 126 (49.2\%) & 144 (53.9\%)\\
\; Meme Enhances Context & 769 (49.1\%) & 351 (50.9\%) & 165 (46.9\%) & 130 (50.8\%) & 123 (46.1\%)\\
\midrule
\multicolumn{6}{l}{\textbf{Meme Types (MT)}} \\
\; Pure Meme     & 218 (13.9\%) & 101 (14.6\%) & 32 (9.1\%)   & 49 (19.1\%)  & 36 (13.5\%) \\
\; Text-in-Meme  & 834 (53.3\%) & 365 (52.9\%) & 189 (53.7\%) & 133 (52.0\%) & 147 (55.1\%)\\
\; Text-out-Meme & 247 (15.8\%) & 117 (17.0\%)  & 57 (16.2\%)  & 36 (14.1\%)  & 37 (13.9\%)\\
\; Comics         & 175 (11.2\%) & 68 (9.9\%)   & 52 (14.8\%)  & 24 (9.4\%)   & 31 (11.6\%) \\
\; Combination   & 91 (5.8\%)   & 39 (5.6\%)   & 22 (6.3\%)   & 14 (5.5\%)   & 16 (6.0\%)  \\
\midrule
\multicolumn{6}{l}{\textbf{Comment Stance (CS)}} \\
\; Support   & 732 (46.8\%) & 326 (47.2\%) & 162 (46.0\%) & 118 (46.1\%) & 126 (47.2\%)\\
\; Denial    & 248 (15.8\%) & 109 (15.8\%)  & 55 (15.6\%)  & 41 (16.0\%)  & 43 (16.1\%)\\
\; Extension & 585 (37.4\%) & 255 (37.0\%) & 135 (38.4\%) & 97 (37.9\%)  & 98 (36.7\%)\\
\midrule
\multicolumn{6}{l}{\textbf{Comment Affection (CA)}} \\
\; Consistent   & 1194 (76.3\%) & 526 (76.2\%) & 267 (75.9\%) & 196 (76.6\%) & 205 (76.8\%) \\
\; Inconsistent & 371 (23.7\%)  & 164 (23.8\%) & 85 (24.1\%) & 60 (23.4\%) & 62 (23.2\%)    \\
\bottomrule
\end{tabular}}
\caption{Comprehensive statistics of the MemeReaCon Benchmark Dataset showing distribution of all annotation categories across subreddits. Percentages in the ``Total Count'' column represent proportion of each category within its group, while percentages in subreddit columns show the distribution within that subreddit.}
\label{tab:dataset_stats}
\end{table*}

\begin{table*}[!t]
\centering
\adjustbox{max width=\linewidth}{
\begin{tabular}{lrrrrrr}
\toprule
\multirow{2}{*}{Category} & \multicolumn{2}{c}{Post Type} & \multicolumn{2}{c}{Comment Stance} & \multicolumn{2}{c}{Comment Affection} \\
\cmidrule(lr){2-3} \cmidrule(lr){4-5} \cmidrule(lr){6-7}
 & CEM & MEC & Support & Deny/Ext. & Consist. & Inconsist. \\
\midrule
\; Pure Meme & 103 (47.2\%) & 115 (52.8\%) & 97 (44.5\%) & 121 (55.5\%) & 167 (76.6\%) & 51 (23.4\%) \\
\; Text-in-Meme & 431 (51.7\%) & 403 (48.3\%) & 397 (47.6\%) & 437 (52.4\%) & 637 (76.4\%) & 197 (23.6\%) \\
\; Text-out-Meme & 128 (51.8\%) & 119 (48.2\%) & 118 (47.8\%) & 129 (52.2\%) & 189 (76.5\%) & 58 (23.5\%) \\
\; Comics & 87 (49.7\%) & 88 (50.3\%) & 79 (45.1\%) & 96 (54.9\%) & 133 (76.0\%) & 42 (24.0\%) \\
\; Combination & 47 (51.6\%) & 44 (48.4\%) & 41 (45.1\%) & 50 (54.9\%) & 68 (74.7\%) & 23 (25.3\%) \\
\bottomrule
\end{tabular}}
\caption{Cross-category distributions showing how different annotation dimensions relate to each other. Percentages represent row proportions.}
\label{tab:cross_relations}
\end{table*}

\begin{table}[!t]
\centering
\adjustbox{max width=\linewidth}{
\begin{tabular}{lrrrrrr}
\toprule
\multirow{2}{*}{Text} & \multicolumn{3}{c}{Word Count} & \multicolumn{3}{c}{Token Count} \\
\cmidrule(lr){2-4} \cmidrule(lr){5-7}
 & Avg. & Max & Min & Avg. & Max & Min \\
\midrule
Post Title  & 7.8  & 24 & 3 & 10.3 & 54 & 3  \\
Meme Text & 14.3 & 68 & 3 & 19.1 & 91 & 4  \\
Top Comment & 16.5 & 89 & 4 & 21.7 & 112 & 5 \\
\midrule
Connection  & 75.4 & 231 & 25 & 80.4 & 246 & 32 \\
Post Intent   & 16.8 & 92  & 11 & 22.9 & 103 & 14 \\
\bottomrule
\end{tabular}}
\caption{Text length statistics across different components of the MemeReaCon dataset. Measurements include both word count and tokenization using the Qwen2.5-32b-instruct tokenizer for consistent evaluation.}
\label{tab:cross_stats}
\end{table}

Tables \ref{tab:dataset_stats}, \ref{tab:cross_stats}, and \ref{tab:cross_relations} provide comprehensive statistics about our dataset, including distributions across different categories, cross-category relationships, and textual characteristics.

\section{Detailed Implementations}
\label{apd:implementations}

This section details the specific prompts and implementation procedures for each task in our MemeReaCon benchmark. The tasks are designed to systematically evaluate models' abilities to understand contextual memes across different dimensions of complexity. All inferences are conducted through vLLM framework or Huggingface Transformers framework. For BERTScore, we use \texttt{microsoft/deberta-xlarge-mnli} as embedding model.

\subsection{Context-Meme Interplay Classification (CMI-C)}
This fundamental task evaluates whether models can identify the primary relationship between the context (post text) and the meme image.

\paragraph{Task Description:} Models must classify the relationship into one of two categories: (1) Context Explain Meme (CEM): The textual context provides necessary information to understand the meme. (2) Meme Enhance Context (MEC): The meme adds additional meaning or humor to the textual context.

\paragraph{Implementation Details:} (1) \textbf{Unimodal Baselines:} For text-only models, we provide detailed descriptions of the meme images. We summarize the descriptions using gpt-4o model via OpenAI API. For image-only models, we render the post text onto the image as a composite manually. (2) \textbf{VLM Models:} Receive both the post text and meme image directly through their respective modality inputs. (3) \textbf{VRM Models:} Receive the same inputs as VLM models but are additionally instructed to explain their reasoning before providing the final classification.

The prompt is shown in Table~\ref{tab:prompt_CMI}.

\begin{table*}[!t]
    \centering
    \small
    \begin{tabular}{|l|}
    \hline
    \textbf{For VLM and VRM models}\\
    Given this social media post with text and an image meme:\\\\
    
    Post text: <post text>\\
    <Meme image is provided>\\\\
    
    Analyze the relationship between the post text and the meme image. Determine which of the following is true:\\
    A. The post text primarily explains or provides context needed to understand the meme image (CEM).\\
    B. The meme image primarily enhances, illustrates, or adds humor to the post text (MEC).\\\\
    
    Select only A or B.\\\\
    
    \textbf{For text-only models}\\
    Given this social media post with text and an image meme:\\\\
    
    Post text: <post text>\\
    Meme description: <meme description>\\\\
    
    Analyze the relationship between the post text and the meme image. Determine which of the following is true:\\
    A. The post text primarily explains or provides context needed to understand the meme image (CEM).\\
    B. The meme image primarily enhances, illustrates, or adds humor to the post text (MEC).\\\\
    
    Select only A or B.\\\\
    
    \textbf{For image-only models}\\
    <Meme image and post text are provided as a composite image>\\\\
    
    Analyze this social media post. Determine which relation is true:\\
    A. Context Explain Meme (CEM)\\
    B. Meme Enhance Context (MEC)\\
    Select only A or B.\\
    \hline
    \end{tabular}
    \caption{Prompt for Context-Meme Interplay task.}
    \label{tab:prompt_CMI}
\end{table*}

\subsection{Comment Stance and Affective Consistence Classification (CSAC-C)}
This dual-aspect task evaluates models' abilities to analyze social dynamics in comments related to meme posts.

\paragraph{Task Description:} Models must: (1) Determine the stance of a comment relative to the original post (support, deny, or extension). (2) Detect whether the comment's literal meaning matches its intended meaning (consistent and inconsistent).

\paragraph{Implementation Details:}
\begin{itemize}
    \item \textbf{Unimodal Baselines:} Similar adaptations as in the CMI-C task, with comment text included.
    \item \textbf{VLM Models:} Process the entire post-meme-comment triple as a unified input.
    \item \textbf{VRM Models:} Are additionally prompted to consider social and cultural contexts that might influence interpretation of stance and affection.
\end{itemize}

\paragraph{Evaluation Metrics:} Accuracy and macro F1-score for the combined classification task with the following matrix (Table~\ref{tab:comment_matrix}):

\begin{table}[!t]
    \centering
    \adjustbox{max width=\linewidth}{
    \begin{tabular}{llll}
        \toprule
        & \textbf{Support} & \textbf{Deny} & \textbf{Extension}\\
        \midrule
        \textbf{Consistent} & Support & Deny & Extension\\
        \textbf{Inconsistent} & Deny & Support & Extension\\
        \bottomrule
    \end{tabular}}
    \caption{Real comments type matrix to show both literal meaning and intended meaning.}
    \label{tab:comment_matrix}
\end{table}

\begin{table*}[!t]
    \centering
    \small
    \begin{tabular}{|l|}
    \hline
    \textbf{For VLM and VRM models}\\
    Analyze this social media interaction:\\\\
    
    Post text: <post text>\\
    <Meme image is provided>\\
    Comments: <comment>\\\\
    
    Part 1 - Stance Analysis: Determine the stance of the comment toward the post:\\
    A. Support (the comment agrees with or reinforces the post)\\
    B. Deny (the comment disagrees with or contradicts the post)\\
    C. Extension (the comment adds information without clearly agreeing or disagreeing)\\\\
    
    Part 2 - Affection Analysis: Determine whether:\\
    A. Consistent (the comment means exactly what it says)\\
    B. Inconsistent (the comment uses irony, sarcasm, or other figurative language)\\\\
    
    Provide your answer as two letters, one for each part (e.g., "A, B").\\\\
    
    \textbf{For text-only models}\\
    Analyze this social media interaction:\\\\
    
    Post text: <post text>\\
    Meme description: <meme description>\\
    Comments: <comment>\\\\
    
    Part 1 - Stance Analysis: Determine the stance of the comment toward the post:\\
    A. Support (the comment agrees with or reinforces the post)\\
    B. Deny (the comment disagrees with or contradicts the post)\\
    C. Extension (the comment adds information without clearly agreeing or disagreeing)\\\\
    
    Part 2 - Affection Analysis: Determine whether:\\
    A. Consistent (the comment means exactly what it says)\\
    B. Inconsistent (the comment uses irony, sarcasm, or other figurative language)\\\\
    
    Provide your answer as two letters, one for each part (e.g., "A, B").\\\\
    
    \textbf{For image-only models}\\
    <Meme image, post text, and comments are provided as a composite image>\\\\
    
    Analyze this interaction. Determine:\\
    1. Comment stance: A. Support, B. Deny, C. Extension\\
    2. Comment tone: A. Consistent, B. Inconsistent\\
    Answer with two letters (e.g., "A, B").\\
    \hline
    \end{tabular}
    \caption{Prompt for Comment Stance + Affection task.}
    \label{tab:prompt_CSCA}
\end{table*}

The prompt is shown in Table~\ref{tab:prompt_CSCA}.

\subsection{Post Connections Generation (PC-G)}
This generative task evaluates models' abilities to articulate the relationships between all elements of a meme post.

\paragraph{Task Description:} Models must generate a free-form explanation that demonstrates understanding of how the post text, meme image, and comments interrelate.

\paragraph{Implementation Details:} All models receive adapted inputs as described in previous tasks. The prompt is shown in Table~\ref{tab:prompt_PC}.

\begin{table*}[!t]
    \centering
    \small
    \begin{tabular}{|l|}
    \hline
    \textbf{For VLM and VRM models}\\
    Analyze this social media post and its comments:\\\\
    
    Post text: <post text>\\
    <Meme image is provided>\\
    Comments: <comment>\\\\
    
    Explain in 3-5 sentences how the following elements connect and interact:\\
    1. The relationship between the post text and the meme image\\
    2. How the comments respond to the post's message\\
    3. Whether the post achieves its apparent communicative purpose\\\\
    
    Be specific about how visual and textual elements work together to create meaning.\\\\
    
    \textbf{For text-only models}\\
    Analyze this social media post and its comments:\\\\
    
    Post text: <post text>\\
    Meme description: <meme description>\\
    Comments: <comment>\\\\
    
    Explain in 3-5 sentences how the following elements connect and interact:\\
    1. The relationship between the post text and the meme image\\
    2. How the comments respond to the post's message\\
    3. Whether the post achieves its apparent communicative purpose\\\\
    
    Be specific about how visual and textual elements work together to create meaning.\\\\
    
    \textbf{For image-only models}\\
    <Meme image, post text, and comments are provided as a composite image>\\\\
    Explain how the text, image, and comments in this post connect. Focus on:\\
    1. Text-image relationship\\
    2. Comment responses\\
    3. Post effectiveness\\
    (3-5 sentences)\\
    \hline
    \end{tabular}
    \caption{Prompt for Post Connection task.}
    \label{tab:prompt_PC}
\end{table*}

\subsection{Post Intent Generation (PI-G)}
This advanced task tests models' abilities to infer the implicit communicative intent behind meme posts.

\paragraph{Task Description:} Models must identify the poster's likely intent, and generate with free-form sentence to show the specific author's intent.

\paragraph{Implementation Details:} All models receive adapted inputs as described in previous tasks. The prompt is shown in Table~\ref{tab:prompt_PIP}.

\begin{table*}[!t]
    \centering
    \small
    \begin{tabular}{|l|}
    \hline
    \textbf{For VLM and VRM models}\\
    Analyze this social media post with its meme and comments:\\\\
    
    Post text: <post text>\\
    <Meme image is provided>\\
    Comments: <comment>\\\\
    
    Based on all available evidence, what was the poster's primary communicative intent?\\
    The intent means the purpose, aim, or goal behind an action, statement, or piece of communication.\\
    It represents what a person or entity intends to convey or achieve.\\
    Provide your answer as a brief sentence.\\\\
    
    \textbf{For text-only models}\\
    Analyze this social media post with its meme and comments:\\\\
    
    Post text: <post text>\\
    Meme description: <meme description>\\
    Comments: <comment>\\\\
    
    Based on all available evidence, what was the poster's primary communicative intent?\\
    The intent means the purpose, aim, or goal behind an action, statement, or piece of communication.\\
    It represents what a person or entity intends to convey or achieve.\\
    Provide your answer as a brief sentence.\\\\
    
    \textbf{For image-only models}\\
    <Meme image, post text, and comments are provided as a composite image>\\\\
    
    What was the poster's primary communicative intent?\\
    The intent means the purpose, aim, or goal behind an action, statement, or piece of communication.\\
    It represents what a person or entity intends to convey or achieve.\\
    Provide your answer as a brief sentence.\\
    \hline
    \end{tabular}
    \caption{Prompt for Post Intent Prediction task.}
    \label{tab:prompt_PIP}
\end{table*}

\section{Further Analysis Experiments}

\subsection{Community-Specific Performance Analysis}
\label{apd:community_specific_performance_analysis}

\begin{table*}[!t]
\centering
\adjustbox{max width=\linewidth}{
\begin{tabular}{lccccc}
\toprule
Model & r/memes & r/meme & r/ProgrammerHumor & r/BritishMemes & r/RelationshipMemes \\
\midrule
\multicolumn{6}{l}{\textit{CMI-C Task (Accuracy \%)}} \\
Qwen2.5-VL & 64.57 & 65.12 & 53.28 & 58.76 & 65.79 \\
InternVL3 & 61.32 & 63.46 & 51.43 & 57.21 & 62.94 \\
Gemini-2.5-pro & 85.72 & 86.19 & 72.33 & 77.83 & 88.03 \\
\hdashline
Max $\Delta$ & 24.40 & 22.73 & 20.90 & 20.62 & 25.09 \\
\midrule
\multicolumn{6}{l}{\textit{PI-G Task (ROUGE-L \%)}} \\
Qwen2.5-VL & 18.34 & 19.07 & 12.52 & 14.79 & 17.63 \\
InternVL3 & 15.87 & 16.92 & 10.28 & 12.44 & 15.97 \\
Gemini-2.5-pro & 45.41 & 46.13 & 35.04 & 39.54 & 45.59 \\
\hdashline
Max $\Delta$ & 29.54 & 29.21 & 24.76 & 27.10 & 29.62 \\
\bottomrule
\end{tabular}}
\caption{Performance across subreddits for representative models. Best and worst performance for each model are highlighted. Max $\Delta$ shows the gap between highest and lowest performing models.}
\label{tab:subreddit_performance}
\end{table*}

Understanding how models perform across different online communities provides critical insights into their ability to comprehend diverse social contexts. We analyze model performance across five popular subreddits to assess how community-specific knowledge affects contextual understanding capabilities.

Table~\ref{tab:subreddit_performance} reveals a consistent and significant performance drop across all models when processing content from specialized communities. All evaluated models perform substantially worse on r/ProgrammerHumor (requiring technical knowledge) and r/BritishMemes (requiring cultural context) compared to general meme communities. Interestingly, we observe that the performance gap between specialized and general communities widens as task complexity increases. For the generative PI-G task requiring deeper contextual reasoning, performance degradation is more severe than for the classification-based CMI-C task. This suggests that specialized knowledge gaps compound when models must perform multi-step reasoning, revealing a fundamental limitation in current contextual understanding capabilities.

The consistent performance differential across community types persists regardless of model scale or architecture, indicating that current pre-training approaches may not adequately capture the specialized knowledge and cultural contexts necessary for understanding community-specific content. This finding challenges the assumption that scaling alone can solve contextual understanding problems, suggesting that targeted approaches to incorporate domain-specific knowledge may be necessary for developing models with robust cross-community understanding capabilities.

\subsection{Meme Structure Performance Analysis}
\label{apd:meme_structure_analysis}

\begin{table}[!t]
\centering
\adjustbox{max width=\linewidth}{
\begin{tabular}{lccccc}
\toprule
Model Type & PM & TIM & TOM & Comic & Comb \\
\midrule
Qwen-VL & 58.73 & 64.92 & 60.37 & 54.68 & 52.94 \\
InternVL3 & 65.28 & 71.43 & 67.58 & 59.76 & 59.05 \\
\hdashline
VLMs (avg) & 58.42 & 65.31 & 60.73 & 55.14 & 52.87 \\
VRMs (avg) & 68.73 & 80.18 & 73.42 & 64.52 & 65.76 \\
\hdashline
$\Delta^*$ & 10.31 & 14.87 & 12.69 & 9.38 & 12.89 \\
\bottomrule
\end{tabular}}
\caption{Impact of meme structural configuration on PC-G task performance. PM: Pure Memes without text overlay; TIM: Text-in-Meme; TOM: Text-out-Meme; Comic: comic format; Comb: Multiple images combination. $\Delta^*$ indicates average performance gap between VRMs and VLMs.}
\label{tab:meme_composition}
\end{table}

The structural configuration of memes significantly impacts model comprehension, revealing important insights about how LVLMs process multimodal content. Table~\ref{tab:meme_composition} shows performance across five distinct meme structures: pure meme (PM), Text-in-Meme (TIM), Text-out-Meme (TOM), comics (Comic), and combination (Comb).

Our analysis reveals a consistent pattern where Vision Reasoning Models (VRMs) substantially outperform standard Vision-Language Models (VLMs) across all structural configurations, with an average performance gap of 10-15\%. This gap widens most dramatically for Text-in-Meme (TIM, $\Delta=14.87\%$), suggesting that VRMs possess superior capabilities for integrating visual and textual elements when they spatially overlap. Interestingly, all models struggle most with comic formats and combination formats (Comb), which require tracking narrative flow across sequential images and understanding relationships between multiple visual elements.

The performance hierarchy (TIM > TOM > PM > Comic > Comb) across model types indicates that current architectures find it easier to process memes where text and image are tightly integrated in a single visual space, compared to formats requiring sequential reasoning or cross-referencing between multiple visual elements. This finding highlights a critical limitation in current LVLMs: while they can effectively process localized multimodal information, they struggle with distributed multimodal reasoning tasks that more closely resemble how humans process complex social media content. The substantial performance degradation on combined formats (12.89\% below TIM for VRMs) demonstrates that even state-of-the-art models have not yet bridged the gap between processing isolated multimodal elements and understanding holistic multimodal narratives.

\subsection{Meme Text-Density Analysis}
\label{apd:meme_text-density_analysis}

\begin{figure*}[!t]
\centering
\includegraphics[width=\linewidth]{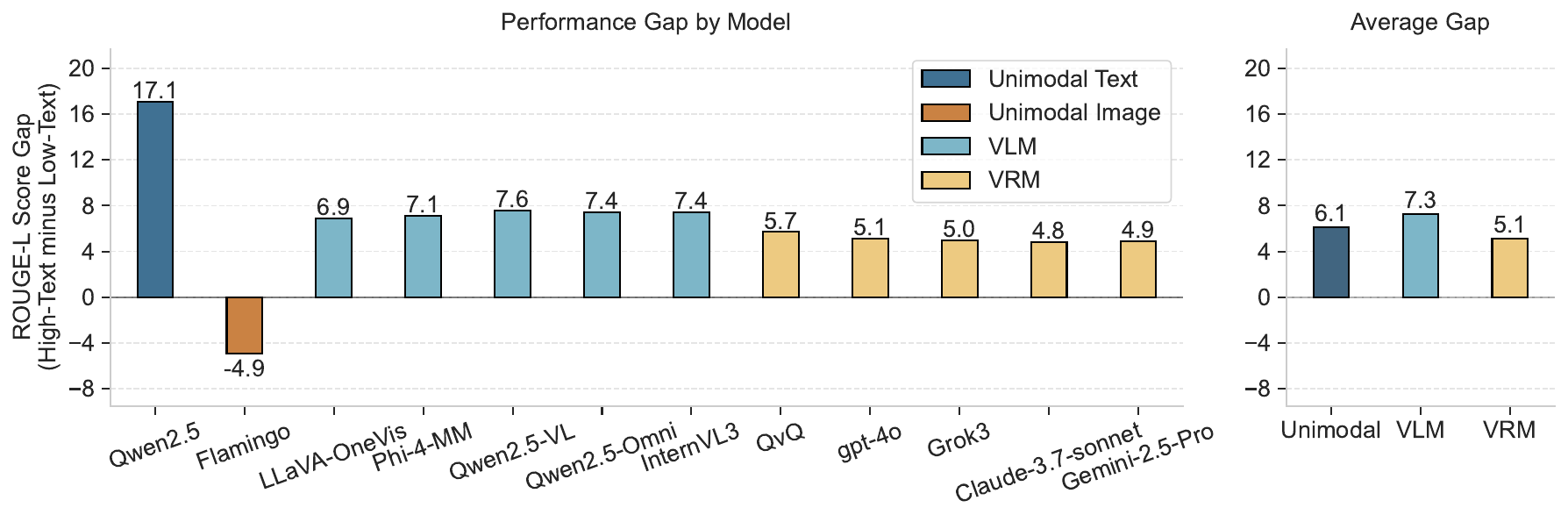}
\caption{Performance gap between high-text and low-text memes across model categories. Positive values indicate better performance on high-text memes. The gap narrows significantly for Vision Reasoning Models, demonstrating their superior cross-modal integration capabilities.}
\label{fig:text_density_gap}
\end{figure*}

Memes exhibit significant variation in text density, ranging from image-dominant formats with minimal text to text-heavy variants where the visual component serves primarily as a backdrop. This variability presents unique challenges for multimodal understanding. To systematically investigate how text density affects model performance, we categorized memes in our dataset into three distinct groups: low-text (0-10 words), medium-text (11-30 words), and high-text (>30 words). This analysis specifically focuses on the Post Intent Prediction (PI-G) tasks, as this requires comprehensive integration of visual and textual elements.

As illustrated in Figure~\ref{fig:text_density_gap}, the performance gap between high-text and low-text memes narrows significantly as model sophistication increases. While VLMs show an average ROUGE-L performance difference of 7.3 points between high-text and low-text memes, this gap shrinks to just 4.7 points for VRMs. Claude-3.7-sonnet exhibits the smallest gap at 4.8 points, suggesting that advanced reasoning mechanisms enable more balanced processing of multimodal content regardless of text-image ratio. This finding has significant implications for meme understanding systems, indicating that sophisticated reasoning capabilities, rather than simply larger model size, are crucial for handling the diverse spectrum of meme formats encountered in real-world social media.

\subsection{Comment Affection Analysis}
\label{apd:comment_affection_analysis}

\begin{table*}[!t]
\centering
\adjustbox{max width=\linewidth}{
\begin{tabular}{lccccccc}
\toprule
\multirow{2}{*}{\textbf{Model}} & \multirow{2}{*}{\textbf{No Comments}} & \multicolumn{3}{c}{\textbf{Consistent Affection}} & \multicolumn{3}{c}{\textbf{Inconsistent Affection}} \\
\cmidrule(lr){3-5} \cmidrule(lr){6-8}
& & Support & Deny & Extend & Support & Deny & Extend \\
\midrule
Qwen2.5-VL & 66.2 & 62.3 & 58.7 & 55.2 & 40.1 & 37.2 & 43.8 \\
Gemini-2.5-Pro & 83.2 & 82.7 & 76.1 & 73.8 & 58.2 & 52.0 & 61.3 \\
\bottomrule
\end{tabular}}
\caption{Model performance on Post Intent Prediction (PI-G) task with different comment types and affection patterns. Results show ROUGE-L (\%).}
\label{tab:comment_affection}
\end{table*}

Social media conversations often involve complex dynamics where comments may support, deny, or extend the original post while conveying affective meanings that can be inconsistent with their literal content. This section explores how these comment characteristics influence models' ability to understand the relationship between posts and memes.

We designed experiments to analyze how models' performance varies across different comment types (support, deny, extension) and affection patterns (consistent vs. inconsistent). Consistent affection occurs when the literal meaning aligns with the intended sentiment (e.g., sincere praise), while inconsistent affection involves misalignment (e.g., sarcastic ``praise'' that actually criticizes). We present data to models under three conditions: (1) without comments, (2) with consistent-affection comments, and (3) with inconsistent-affection comments. For each condition, we evaluated performance on the Post Intent Prediction (PI-G) task, which requires inferring the poster's communicative intent.

As shown in Table~\ref{tab:comment_affection}, both Gemini-2.5-pro and Qwen2.5-VL models experience a substantial performance disparity between consistent and inconsistent affection scenarios. When presented with comments whose affective meaning contradicts their literal content (inconsistent affection), even leading Vision Reasoning Models (VRMs) suffer performance drops of 20-25 percentage points compared to consistent affection scenarios. This gap, which we term the ``Context-Affection Gap,'' is most pronounced in deny comments with inconsistent affection (e.g., sarcastic agreement that actually contradicts). For instance, Gemini-2.5-Pro achieves 76.1\% accuracy with consistent denial comments but only 52.0\% with inconsistent denial comments.

This finding suggests that current LVLMs struggle with communication where literal meaning diverges from intended meaning. The narrower gap observed in VRMs compared to VLMs indicates that advanced reasoning models are hurt more by providing opposite points of view.

\subsection{Modality Contribution Analysis}
\label{apd:modality_analysis}

\begin{table*}[!t]
\centering
\adjustbox{max width=\linewidth}{
\begin{tabular}{lcccccccc}
\toprule
\multirow{2}{*}{Settings} & \multicolumn{2}{c}{PC-G} & \multicolumn{2}{c}{PI-G} & \multicolumn{2}{c}{PC-G} & \multicolumn{2}{c}{PI-G} \\
\cmidrule(lr){2-3} \cmidrule(lr){4-5} \cmidrule(lr){6-7} \cmidrule(lr){8-9}
& R-L (\%) & $\Delta$ & R-L (\%) & $\Delta$ & R-L (\%) & $\Delta$ & R-L (\%) & $\Delta$\\
\midrule
&\multicolumn{4}{c}{\textbf{\textit{Qwen2.5-VL}}}&\multicolumn{4}{c}{\textbf{\textit{Gemini-2.5-Pro}}}\\
Original     & 60.38 & -      & 44.86 & -      & 38.92 & -      & 20.43 & -     \\
\hdashline
No Text      & 48.25 & -12.13 & 36.42 & -8.44  & 30.47 & -8.45  & 10.69 & -9.74 \\
Random Text  & 52.81 & -7.57  & 38.67 & -6.19  & 32.50 & -6.42  & 15.30 & -5.13 \\
\hdashline
No Image     & 26.10 & -34.28 & 21.69 & -23.17 & 22.66 & -16.26 & 14.98 & -5.45 \\
Random Image & 28.34 & -31.84 & 23.45 & -21.21 & 26.88 & -12.04 & 17.25 & -3.18 \\
\bottomrule
\end{tabular}}
\caption{Performance of modality contribution analysis. ``Original'' uses the meme's actual context; ``Random Text'' and ``Random Image'' uses mismatched context/image from a different post. ``No Text'' and ``No Image'' removes post title/image. Text modified for Meme to Enhance Context setting (MEC), while image modified for Context to explain meme setting (CEM).}
\label{tab:context_relevance}
\end{table*}

To investigate how different elements of posts contribute to model understanding, we conducted systematic ablation experiments by removing or replacing key components. Table~\ref{tab:context_relevance} shows performance changes when manipulating either textual context or visual elements.

Our findings reveal several interesting patterns. First, image removal causes dramatically larger performance drops than text removal, with PC-G task performance declining by 34.28\% for Qwen2.5-VL compared to just 12.13\% when text is removed. This suggests that memes serve as the primary information carrier in these multimodal posts, even for the ``Meme to enhance context'' setting. Second, models perform better with mismatched components than with missing ones: random text produces smaller drops (7.57\% for Qwen2.5-VL on PC-G) than no text (12.13\%). This indicates models use whatever context is available to create meaning, even when connections are tenuous.

Most surprisingly, we find that smaller models like Qwen2.5-VL show greater sensitivity to modality manipulation than larger ones like Gemini-2.5-Pro. When presented with random images, Qwen2.5-VL's performance drops by 31.84\% on PC-G tasks, while Gemini-2.5-Pro decreases by only 12.04\%. This suggests that reasoning models develop more robust internal representations that can partially recover from contextual mismatches, effectively ``filtering out'' irrelevant information. These findings highlight a critical gap in current models: while they can process multimodal inputs, they struggle to determine which elements should be contextually emphasized or disregarded, which is a fundamental aspect of human social media consumption that remains challenging for LVLMs.

\section{Error Analysis Description and Performance}
\label{apd:error}

We categorized errors into four distinct patterns that emerged consistently across models (Figure~\ref{fig:error_categories}). The distribution of these errors varies significantly between model architectures, revealing fundamental differences in contextual processing capabilities.

The four primary error patterns we identified are:

\begin{figure}[!t]
\centering
\includegraphics[width=\linewidth]{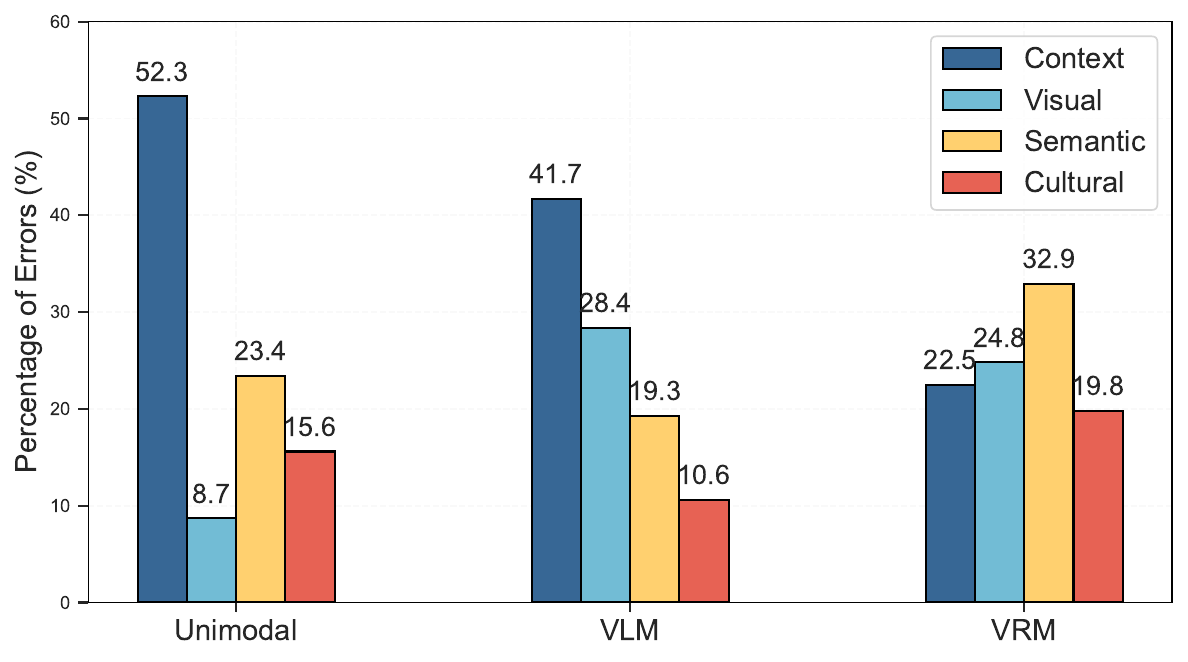}
\caption{Distribution of error types across model categories when interpreting memes in context. Vision Reasoning Models (VRMs) make fewer context-neglect errors but struggle more with contextual conflicts than Vision-Language Models (VLMs).}
\label{fig:error_categories}
\end{figure}

\begin{itemize}[leftmargin=*, itemsep=0pt]
    \item \textbf{Context:} Models process the meme in isolation, disregarding crucial context from the post text or comments. This was most prevalent in VLMs (41.7\%) and less common in VRMs (22.5\%), suggesting that reasoning-enhanced architectures better incorporate textual context.
    \item \textbf{Visual:} Models overemphasize visually important but contextually irrelevant image elements. This error occurred when models focused on character objects rather than the socially relevant aspects indicated by the post.
    \item \textbf{Semantic:} Initially correct interpretations gradually go wrong as response length increases. Notably, this was highest among VRMs (32.9\%), suggesting that more powerful generative capabilities sometimes lead to unfocused elaboration.
    \item \textbf{Cultural:} Models fail to recognize community-specific references, slang, or humor conventions. This affects all model classes but was most pronounced in VRMs (19.8\%), possibly due to their attempts at more complex reasoning about unfamiliar cultural elements.
\end{itemize}

\section{More Error Cases}
\label{apd:more_error_cases}

\begin{figure*}[!t]
    \centering
    \includegraphics[width=\linewidth]{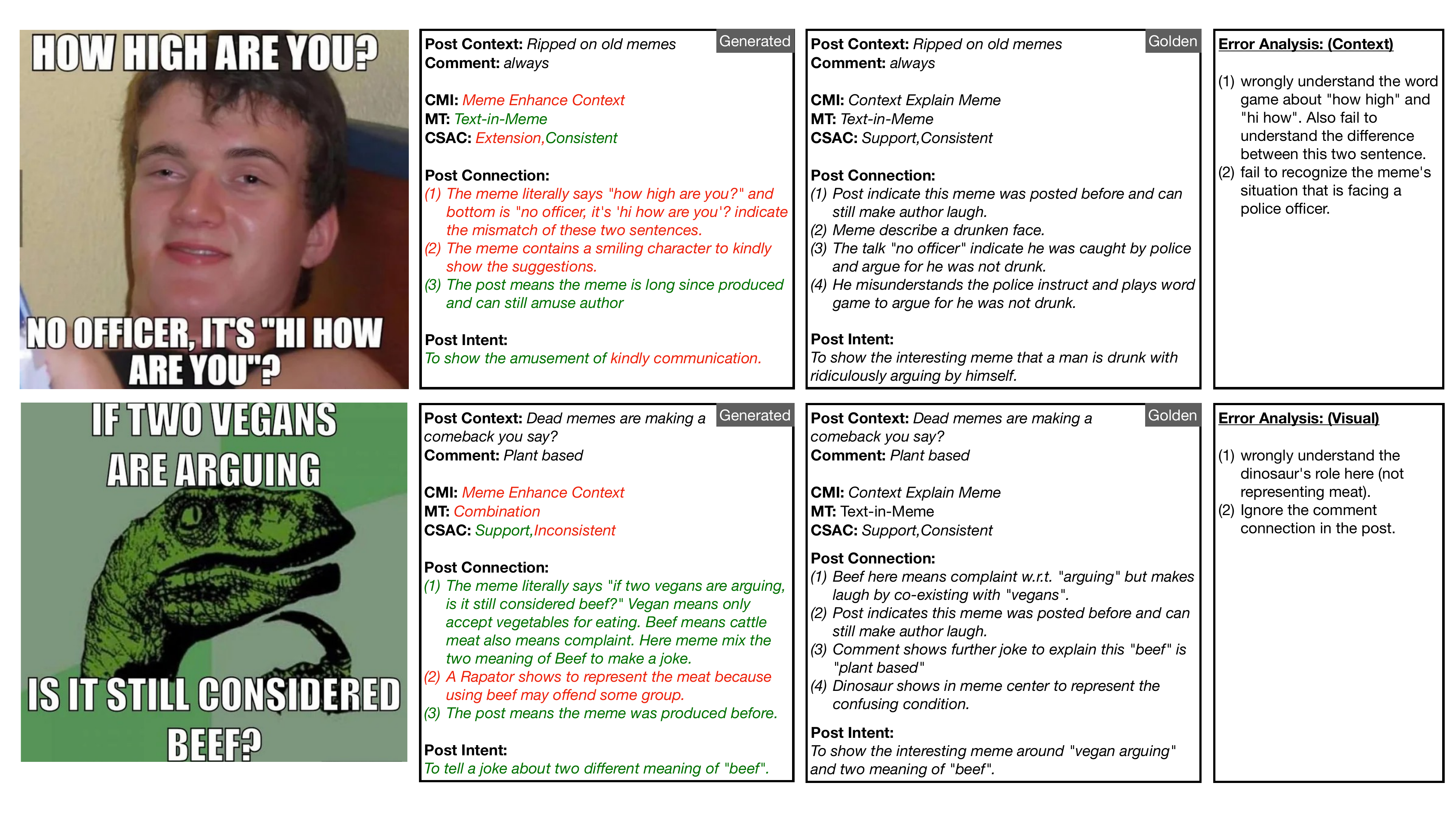}
    \caption{Error cases for Context error and Visual error. The green text indicates the correct answer compared with golden. The red text indicates the wrong answer.}
    \label{fig:more_error_cases_1}
\end{figure*}

\begin{figure*}[!t]
    \centering
    \includegraphics[width=\linewidth]{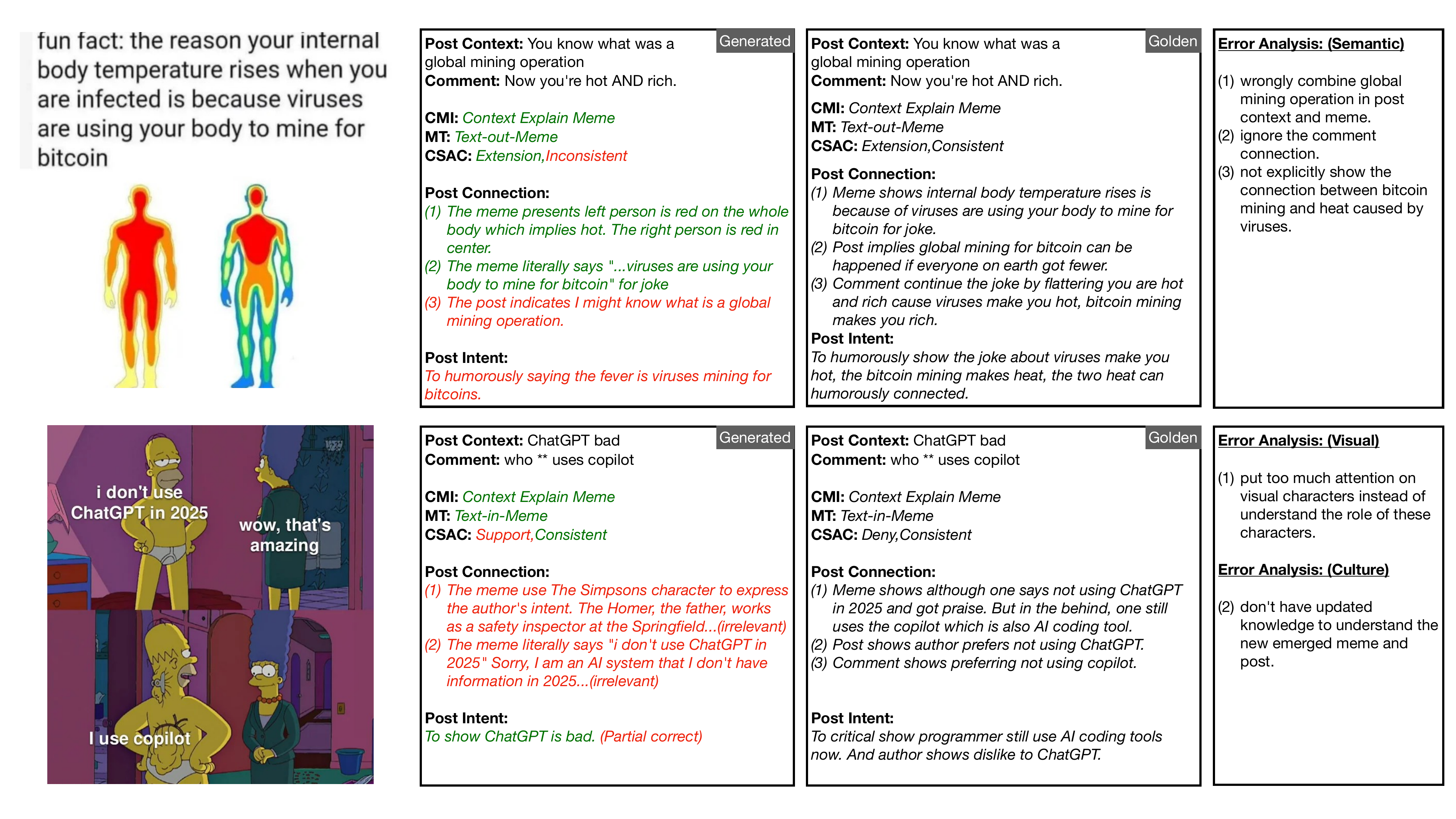}
    \caption{Error cases for Semantic error and Culture error. The green text indicates the correct answer compared with golden. The red text indicates the wrong answer.}
    \label{fig:more_error_cases_2}
\end{figure*}

We show more error cases covering each error type in Figure~\ref{fig:more_error_cases_1} and~\ref{fig:more_error_cases_2}.

\end{document}